\documentclass[sn-mathphys-num]{sn-jnl}

\usepackage{graphicx}%
\usepackage{multirow}%
\usepackage{amsmath,amssymb,amsfonts}%
\usepackage{amsthm}%
\usepackage{mathrsfs}%
\usepackage[title]{appendix}%
\usepackage{xcolor}%
\usepackage{textcomp}%
\usepackage{manyfoot}%
\usepackage{booktabs}%
\usepackage{algorithm}%
\usepackage{algorithmicx}%
\usepackage{algpseudocode}%
\usepackage{listings}%

\theoremstyle{thmstyleone}%
\theoremstyle{thmstyletwo}%

\theoremstyle{thmstylethree}%

\begin{document}

\title[Article Title]{Large Language Models are Few-shot Multivariate Time Series Classifiers}


\author[1,3]{\fnm{Yakun} \sur{Chen}}

\author[1]{\fnm{Zihao} \sur{Li}}

\author[2]{\fnm{Chao} \sur{Yang}}

\author*[1]{\fnm{Xianzhi} \sur{Wang}}

\author[1,3]{\fnm{Guandong} \sur{Xu}}

\affil*[1]{\orgdiv{School of Computer Science, Faculty of Engineering and Information Techology}, \orgname{University of Technology Sydney}, \orgaddress{\street{15 Broadway}, \city{Sydney}, \postcode{2007}, \state{NSW}, \country{Australia}}}

\affil[2]{\orgdiv{Faculty of Information Technology and Engineering}, \orgname{Ocean University of China}, \orgaddress{\street{Street}, \city{City}, \postcode{10587}, \state{State}, \country{Country}}}

\affil[3]{\orgdiv{Centre for Learning, Teaching and Technology}, \orgname{The Education University of HongKong}, \orgaddress{\street{Street}, \city{City}, \postcode{610101}, \state{State}, \country{Country}}}


\abstract{Large Language Models (LLMs) have been extensively applied in time series analysis. Yet, their utility in the few-shot classification (i.e., a crucial training scenario due to the limited training data available in industrial applications) concerning multivariate time series data remains underexplored.
We aim to leverage the extensive pre-trained knowledge in LLMs to overcome the data scarcity problem within multivariate time series.
Specifically, we propose \textbf{LLMFew}, an LLM-enhanced framework to investigate the feasibility and capacity of LLMs for few-shot multivariate time series classification.
This model introduces a \textbf{P}atch-wise \textbf{T}emporal \textbf{C}onvolution \textbf{Enc}oder (PTCEnc) to align time series data with the textual embedding input of LLMs. We further fine-tune the pre-trained LLM decoder with Low-rank Adaptations (LoRA) to enhance its feature representation learning ability in time series data. 
Experimental results show that our model outperformed state-of-the-art baselines by a large margin, achieving 125.2\% and 50.2\% improvement in classification accuracy on Handwriting and EthanolConcentration datasets, respectively.
Moreover, our experimental results demonstrate that LLM-based methods perform well across a variety of datasets in few-shot MTSC, delivering reliable results compared to traditional models. This success paves the way for their deployment in industrial environments where data are limited. Our code is available at: https://anonymous.4open.science/r/llm-fewshot-mtsc-C608.}




\maketitle

\section{Introduction}


Multivariate Time Series Classification (MTSC) is a challenging task due to its high-dimensional and complex data structures, which requires the consideration of intra-class similarities and inter-class differences across various classes~\cite{IsmailFawaz2018deep}. 
The inherent complexity stems from the need to model and interpret intricate interactions and dependencies among multiple time-indexed variables, which can vary widely in behavior and influence across different domains.
In many scenarios, collecting labeled data to train deep learning models is challenging, leading to the emergence of the few-shot learning problem.
For example, in the healthcare field~\cite{rajkomar2018scalable}, Electrocardiogram (ECG) time series data are important for diagnosing cardiac arrhythmias; however, the labeled ECG data available for training are often limited due to expensive data collection, time-consuming labor annotation, and patient privacy issues.
The data scarcity makes few-shot learning an increasingly important topic.
Another important application is the Internet of Things (IoT) and edge computing, the smart devices make intelligent decisions using their own collected time series data which can also be considered a few-shot learning scenario~\cite{jiang2020decentralized}.

\begin{figure}[t]
\centering
\includegraphics[width=0.60\columnwidth]{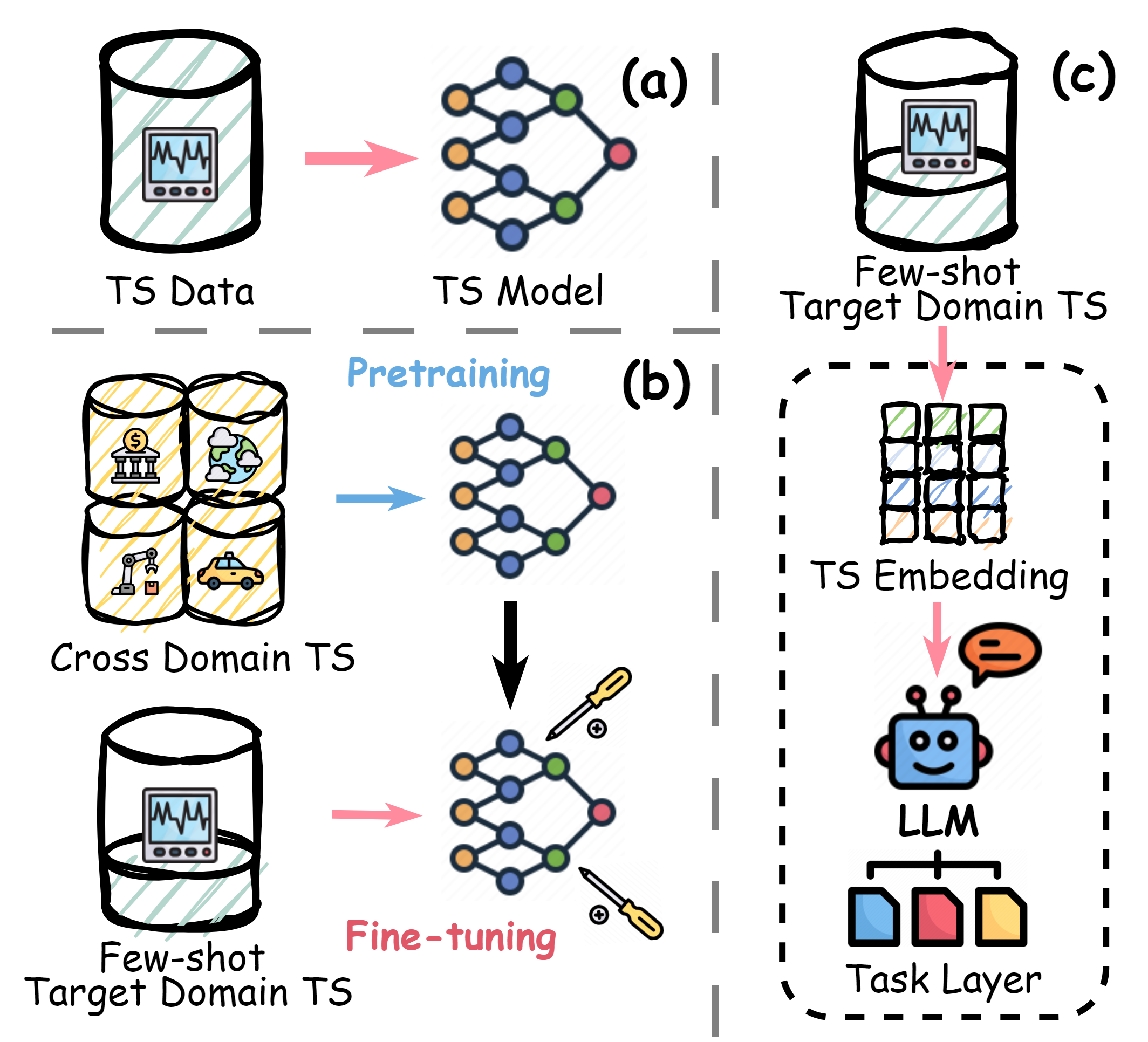} 
\caption{
Illustration of different training paradigms for MTSC. \textbf{(a)} classical end-to-end deep learning, not designed for few-shot scenarios, requires a large amount of labeled training samples; \textbf{(b)} transfer learning uses a pre-training stage to learn knowledge from source time series data from different domains and then fine-tunes the learned model for the target domain with limited training samples; \textbf{(c)} our proposed LLM-enhanced framework for few-shot scenarios without leveraging source time series data.}
\label{fig:intro}
\end{figure}

Classical deep learning models, illustrated in Figure~\ref{fig:intro} (a), tend to overfit small-scale training datasets; they often fail to generalize and cover various test data distributions, causing significant performance degradation under limited training samples.
Existing limited studies on few-shot MTSC~\cite{fawaz2018transfer,zhang2022self} heavily rely on the transfer learning strategy, illustrated in Figure~\ref{fig:intro} (b). These methods generally pre-train a model on various source datasets and then fine-tune it on the target dataset.
However, time series data is domain- and task-sensitive with varying periodicity and patterns, hindering deep learning systems from fitting the distribution gaps across different datasets. Besides, expensive data collection and privacy issues add to the challenges for obtaining large-scale labeled datasets for pre-training or target datasets for transfer learning in many scenarios like healthcare and the Internet of Things~\cite{tang2020interpretable}.
Meta-learning uses a gradient-based optimizer to accelerate model parameter adaptation; it is applied to mitigate the source-target data distribution drift issue~\cite{narwariya2020meta}.
However, all the above methods face challenges in handling the distribution discrepancy between source and target domains~\cite{wang2020generalizing}---the data scarcity problem in the target dataset makes it more difficult to estimate the distribution.


Nowadays, large language models (LLMs) have been applied to time series analysis due to their powerful ability to model sequential dependencies, extending their effectiveness from text to numerical time series data~\cite{jin2024position}.
The current LLM-based methods for time series tasks focus on designing alignment modules and adapters to integrate time series knowledge into LLMs. While this focus helps improve the understanding of temporal patterns, it neglects the ability to utilize pre-trained knowledge to handle few-shot learning scenarios. Furthermore, it remains undetermined whether the success of LLMs in the time series domain is primarily driven by LLMs' pre-trained knowledge or by the underlying transformer architecture.

In this paper, we aim to answer whether the pre-trained knowledge embedded in the LLMs can migrate to model numerical time series dependencies in few-shot learning scenarios of multivariate time series classification.
We treat LLMs' knowledge gained from vast pre-trained text corpus as source time series data in the traditional transfer learning paradigm, thus addressing the challenge of obtaining plenty of source time series data. 
%
Following this idea, we propose \textbf{LLMFew}, a novel LLM-based framework for few-shot MTSC.
This approach exploits LLM's inherent ability to model sequences without requiring many source domain datasets, as illustrated in Figure~\ref{fig:intro} (c).
Specifically, our model applies a \textbf{P}atch-wise \textbf{T}emporal \textbf{C}onvolution \textbf{Enc}oder (PTCEnc) to align the time series embedding to the input dimension of the LLM. Then, we fine-tune part of the parameters in the LLM to make it study temporal patterns within limited training data. Last, an MLP-based classification head is employed to predict the most possible class. 
In summary, this paper makes the following main contributions:
\begin{itemize}
    \item To the best of our knowledge, we are the first to explore leveraging LLMs for few-shot MTSC. Even without source datasets from different domains, LLM-based methods can outperform state-of-the-art baselines by a large margin with pre-trained knowledge assistance. 
    \item We proposed \textbf{LLMFew}, which applies a PTCEnc to convert time series data to embeddings suitable for the LLMs and fine-tunes LLMs for the downstream MTSC task. 
    \item Our proposed model achieved satisfactory performance compared to state-of-the-art baselines under different few-shot settings based on our comprehensive experiments on 10 datasets from different domains. We further analyze the application of LLMs to few-shot MTSC and offer insights into possible future directions in the field.
\end{itemize}

\section{Related Work}
\subsection{Few-shot Learning for MTSC}
Statistical and Machine learning methods have first been applied to MTSC tasks, including Dynamic Time Warping (DTW)~\cite{gudmundsson2008support}, a distance calculation function, can compensate for possible confounding offset with some realignment of series and combined with support vector machine to classify multivariate time series data. Random Convolutional Kernel Transform (ROCKET)~\cite{dempster2020rocket}, with many random convolution kernels, is another efficient method without the parameter update during training. However, they are not effective when dealing with high-dimensional data because require using hand-crafted features. 
Classical statistical and machine learning methods like Dynamic Time Warping (DTW)~\cite{gudmundsson2008support} and Random Convolutional Kernel Transform (ROCKET)~\cite{dempster2020rocket} have been used in multivariate time series classification. DTW adjusts time series alignment and, combined with support vector machines, classifies data effectively. ROCKET uses random convolution kernels and does not update parameters during training. However, both methods struggle with high-dimensional data as they depend on hand-crafted features.

With the development of deep learning methods, Recurrent structures, including GRU and LSTM, have first been used to solve the MTSC problem~\cite{karim2017lstm}. Later, Convolutional Neural Network (CNN) was adopted because of its success in the CV field and higher efficiency for parallel computing compared to the recurrent structure~\cite{tang2020rethinking}. Transformer-based models first emerged for NLP and then were later applied to MTSC because they combine the advantage of both recurrent and convolutional models. Dyformer~\cite{yang2024dyformer} uses a hierarchical pooling layer to decompose time series data and gates to process and fuse information among subsequences. Shapeformer~\cite{le2024shapeformer} considers using shapelets to identify similarities within classes. However, when it comes to few-shot scenarios, the above state-of-the-art models possibly cannot avoid overfitting problems and fail to learn generalizable features within and between classes.

Limited research focuses on few-shot MTSC. The initial idea is to enlarge the training datasets using the data from different domains~\cite{gupta2021similarity} or applying data augmentation methods through time and frequency domain~\cite{zhang2023few}. This type of data is inherently unstable. Performance hinges on whether the model's ability to learn within-class similarities through this process is effective in distinguishing unseen test data across multiple classes. Meta-learning is another possible solution for few-shot scenarios by fusing meta-features and using triplet loss to learn the representations within and across classes~\cite{narwariya2020meta,park2023meta}. A similar design uses contrastive learning, which is a self-supervised learning paradigm, to have a 2-stage training: a self-supervised pre-training and label supervised fine-tuning~\cite{chen2023supervised}. However, all the above methods do not actually solve the few-shot problem because they all require external reliable time series data to learn temporal patterns and serve as the knowledge for the downstream classification task.

\subsection{LLM Applications for Time Series Analysis}
LLMs are popular and prominent within the time series analysis, providing a new perspective for learning temporal patterns. Researchers focus on how to reprogram time series data and then feed it into LLMs to improve performance. The first pioneering work~\cite{gruver2024large} converts numerical data into text versions to let the LLMs finish the probabilistic forecasting task. PromptCast considers a similar way to design a hard prompt to encode the time series input~\cite{xue2023promptcast}. However, this kind of method omits the fact that the frequency of numbers appearing in text is very low, and the meaning of numbers in the semantic space is inconsistent with its in the time series data. More recent work considers using alignment modules to convert multivariate time series data into embedding and then put them into LLMs in the next step. Some effective modules in past research include Instance Norm and Patching~\cite{jin2023time,zhou2023one}. Another inspiring direction is decomposing time series into trend, seasonal and residual parts and then adding three parts together to generate input embedding~\cite{cao2023tempo,pan2024s}. All the above efforts are aimed at activating the sequence modeling ability of LLM to understand the temporal features in time series, which are unseen in its pre-training tasks. They also point out the importance of aligning time series and text semantic spaces.

\section{LLMFew}

\subsection{Problem Definition}
\subsubsection{Multivariate Time Series Classification}
Given a multivariate time series data $\mathbf{X} = \{\mathbf{x}_1, \mathbf{x}_2, \cdots, \mathbf{x}_M \} \in \mathbb{R}^{M \times L}$ annotated by a label $y_n \in \mathcal{S}$, $|\mathcal{S}| = N$, where $M$ is the number of series, $L$ is the sequence length, and $N$ is the number of class. The multivariate time series classifier can be formulated as $f:\mathbb{R}^{M \times L} \rightarrow \mathbb{Z}^{N}$, which aims to distinguish the multivariate time series data into $N$ different categories by $f\left(\mathbf{X}\right) = y_n$ and return the class label $y_n$.
\subsubsection{Few-shot Learning (FSL)}
FSL problems are predominantly supervised learning tasks. Specifically, few-shot classification involves developing a classifier based on only a small number of labeled examples for each class.
We consider the N-way-K-shot classification in which the model $f$ is exposed only to $K \times N$ samples. This learning approach is termed as K-shot learning. Let $\mathcal{D}_{train} = {(\mathbf{X}_i, \mathbf{y}_i)}_i^{K \times N}$ be the K-shot training dataset, the objective is to train a model $f$ to minimize the discrepancy between its predictions and the true class labels over the training dataset $\mathcal{D}_{train}$.

\subsection{Framework}

\begin{figure*}[!t]
\centering
\includegraphics[width=0.98\textwidth]{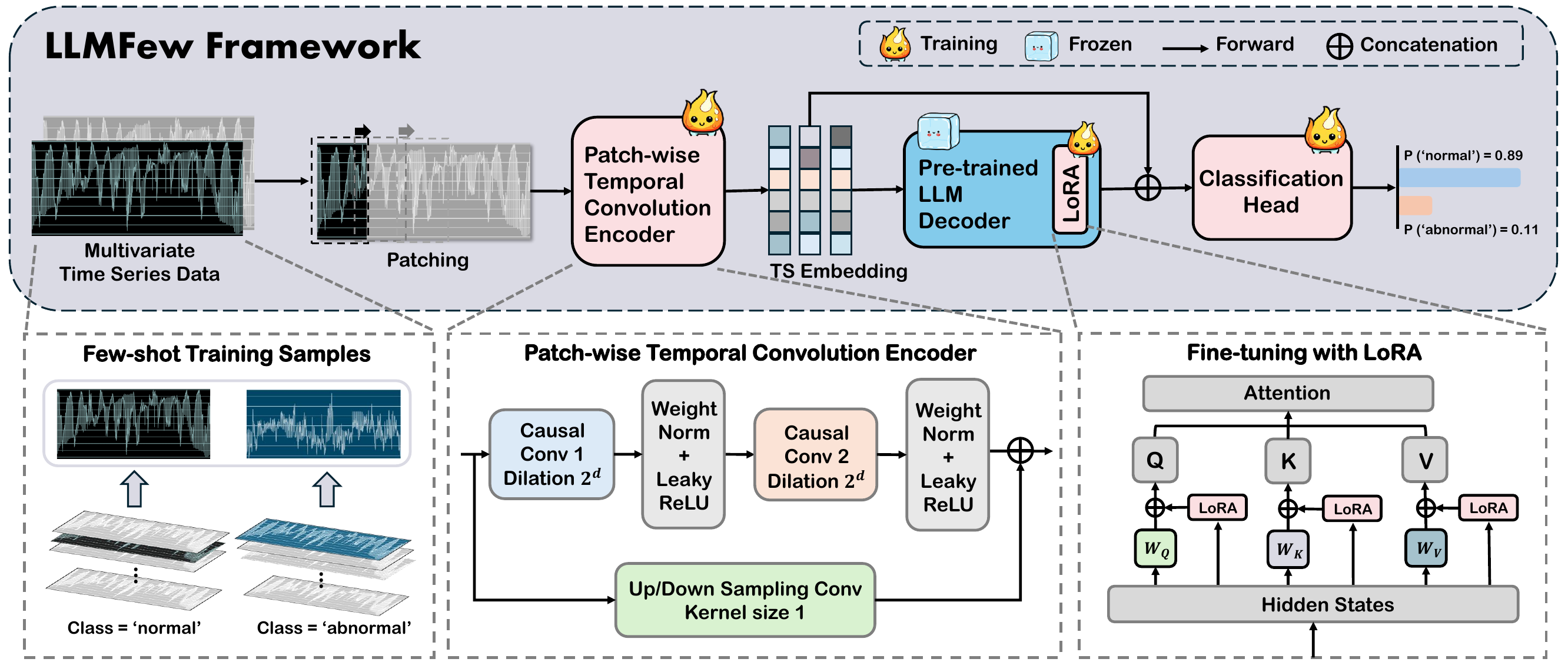}
\caption{
Framework of LLMFew. We first preprocess MTS into a group of patches. The \textbf{P}atch-wise \textbf{T}emporal \textbf{C}onvolution \textbf{Enc}oder (PTCEnc) later converts the multiple patches to the time series embeddings, ready for LLMs. To activate LLMs' ability to learn temporal features, we use LoRA to fine-tune Q, K, V parameters of attention layers. Last, we use a classification head to generate the possible class.}
\label{fig:framework}
\end{figure*}

Our model framework is depicted in Figure~\ref{fig:framework}. LLMFew is an LLM-enhanced framework with four main modules: a Patching layer, a \textbf{P}atch-wise \textbf{T}emporal \textbf{C}onvolution \textbf{Enc}oder (PTCEnc), an LLM decoder with Low-Rank Adaptation (LoRA) fine-tuning and lastly a classification head.

\subsubsection{Patching}
We adopt a Patching Layer to process multivariate time series input, which has been widely used in Transformer-based Models for learning local semantic information~\cite{nie2022time}. The multivariate time series $\mathbf{X}$, with $M$ features and $L$ time steps, can be split into $M$ univariate series $\mathbf{x}^{m} \in \mathbb{R}^{1 \times L}$. Each univariate time series $\mathbf{x}^{m}$ is divided into patches independently. We denote the patch length as $P$ and the stride (the non-overlapping region between two consecutive patches) as $S$, then the patching layer will generate a sequence of patches $\textbf{x}_p^m \in \mathbb{R}^{P \times N_P}$, where $N_P$ is the number of patches, $N_P = \lfloor \frac{(L-P)}{S} \rfloor + 2$. We pad $S$ repeated numbers of the last value $\textbf{x}_L^m \in \mathbb{R}$ to the end of the original sequence before patching to prevent losing information at the end of the sequences. The patching layer can be formalized as:
\begin{equation}
    \mathbf{X}_P = \textrm{Patching}(\mathbf{X})
\end{equation}
where $\mathbf{X}_P \in \mathbb{R}^{M \times P \times N_P}$ are the patches of the input multivariate time series $\mathbf{X}$. After the patching layer, the number of input tokens is reduced from $L$ to approximately $L/S$, which is helpful in reducing the memory usage and computational complexity when fine-tuning the LLM decoder later.

\subsubsection{Patch-wise Temporal Convolution Encoder (PTCEnc)}
We put the patches of the input $\mathbf{X}_P$ into PTCEnc to align the time series representations to the input of LLMs, i.e. text token embeddings. Inspired by the Temporal Convolutional Network (TCN)~\cite{bai2018empirical}, PTCEnc is a structure consisting of stacked causal convolution blocks with depth $D$, which can be formalized as:
\begin{equation}
\begin{split}
    \mathbf{H}^{(d)}_e &= \textrm{ConvBlock}^{(d)}(\mathbf{H}_e^{(d-1)}) +\mathbf{H}_e^{(d-1)} \\
    \textrm{ConvBlock}^{(d)}(\cdot) &= \prod_{i=1}^2 \textrm{CausalLayer}^{(d)}_i (\cdot)  \\
    \textrm{CausalLayer}^{(d)}_i(\cdot) &=  \sigma ( \textrm{Norm}( \textrm{CausalConv}^{(d)}_i(\cdot))) \\
\end{split}
\end{equation}
where $\mathbf{H}^{(0)}_e = \mathbf{X}_P^M$, $\mathbf{H}^{(d)}_e$ is the PTCEnc hidden representation at depth $d$, generated by a skip connection across each $\textrm{ConvBlock}^{(d)}(\cdot)$. Each $\textrm{ConvBlock}^{(d)}(\cdot)$ at depth $d$ consists of two sequential $\textrm{CausalLayer}^{(d)}$ with the same structure. The $\textrm{CausalLayer}^{(d)}$ is implemented by a $\textrm{CausalConv}^{(d)}$ with a dilation factor $2^d$, followed by a weight normalization function $\textrm{Norm}(\cdot)$ and a LeakyReLU activation function $\sigma(\cdot)$.
The final output of PTCEnc $\mathbf{H}_e$ is calculated by $\mathbf{H}_e = \textrm{Conv}(\mathbf{H}^{(D)}_e)$. We use a Sampling $\textrm{Conv}(\cdot)$ to match the dimension of $\mathbf{H}_e$ with LLM's embedding dimension.

\subsubsection{LLM Decoder with LoRA Fine-tuning}
We employ decoder-based LLMs as the backbone to extract time series representations.
Inside the LLM block, we apply LoRA~\cite{hu2021lora}, a Parameter-Efficient Fine-Tuning (PEFT) method to inject trainable rank decomposition matrix $\mathbf{A}$ and $\mathbf{B}$ into each Transformer block in the LLM to reduce the size of trainable parameters. To leverage the sequential ability of LLMs, we apply LoRA to fine-tune the Query, Key and Value parameters of attention layers in LLMs. 
We denote the rank of the LoRA module by $r$, which is a hyperparameter reflecting the amount of information the module can contain. Assuming that the input of Query and Key in an attention layer is $\mathbf{h}_0$ with size $d$ and the output is $\mathbf{h}$ with size $k$, then $\mathbf{A} \in \mathbb{R}^{d \times r}$, $\mathbf{B} \in \mathbb{R}^{r \times k}$, and $r \ll min(d,k)$. This process can be formalized as:
\begin{equation}
    \mathbf{h} = \mathbf{W}_0\mathbf{h}_0 + \frac{\alpha}{r}\mathbf{A}\mathbf{B}\mathbf{h}_0,
\end{equation}

where $\mathbf{W}_0$ denotes the pretrained weight matrix, $\alpha$ is a hyperparameter that acts similarly to the learning rate.
We trained the attention-related parameters in each layer and used the representation of the last hidden state as the output of the LLM decoder. 

\subsubsection{Classification Head and Loss Function}
We perform a skip connection with an element-wise addition of outputs from the PTCEnc and LLM decoder to fuse the two embeddings. 
Lastly, We flattened the fused embedding $\mathbf{H}$ and fed them through the final MLP-based classification head, which outputs a vector with probabilities of each time series class. This process can be formalized as:
\begin{equation}
\begin{split}
    \mathbf{H}_{d} &= \textrm{LLM}(\mathbf{H}_e) \\
    \mathbf{H} &= \textrm{ReLU}(\mathbf{H}_e + \mathbf{H}_{d}) \\
    \mathbf{y} &= \textrm{Softmax}\left(\textrm{LN}(\textrm{Linear}(\mathbf{H})) \right) 
\end{split}
\end{equation}
where $\textrm{LLM}(\cdot)$ can be implemented by any of the LLMs with LoRA fine-tuning, $\textrm{LN}(\cdot)$ is the layer normalization, and $\textrm{Softmax}(\cdot)$ is the softmax activation function.
We use the cross-entropy loss for the classification task, which can be formalized as $\mathcal{L} = - \sum_{n=1}^N\mathbf{y}_n \log(p_n)$, where $N$ is the number of classes, $\mathbf{y}_n$ is a one-hot encoded vector, and $p_i$ is predicted probability that the sample belongs to class $n$.

\section{Experiments}
In our experiments, we first demonstrate that our model consistently outperforms all baselines in $1$-shot learning. 
Second, we discuss model performance in K-shot learning scenarios, with the conclusion that with the increase of k, the model performance increases, but the marginal benefit diminishes.
We also provide the average accuracy when training on the full datasets, and our model can achieve a competitive performance with all baselines.
Besides, the results for different LLMs in our framework show that choosing LLMs is also a data-centric task, especially since Large LLMs do not work for small datasets.
Additionally, we conduct an ablation study to validate the effectiveness of each component in our model. 
We conduct key hyperparameter sensitivity analysis, and their values should be chosen based on the dimension and sequence length of the dataset. Finally, we assessed inference time and memory cost.

\subsection{Experimental Setup}
\subsubsection{Datasets}
Following~\cite{zhou2023one}, we choose 10 MTSC datasets from the UEA Archive~\cite{bagnall2018uea}, covering different fields: EthanolConcentration (EC), FaceDetection (FD), Handwriting (HW), Heartbeat (HB), JapaneseVowels (JV), PEMS-SF (PS), SelfRegulationSCP1 (SCP1), SelfRegulationSCP2 (SCP2), SpokenArabicDigits (SAD), and UWaveGestureLibrary (UGL). These datasets provide different characteristics, with a range of features from 3 to 963, sequence length up to 1751 and the number of classes up to 26. Table~\ref{tab:datasets} shows the statistics of datasets.

\begin{table*}[t]
  \centering
  \setlength{\tabcolsep}{1.5mm} 
  \caption{Datasets Satistics}
    \begin{tabular}{ccrrrrr}
    \toprule
    Abbr. & Datasets & \multicolumn{1}{c}{Train Size} & \multicolumn{1}{c}{Test Size} & \multicolumn{1}{l}{Dimensions} & \multicolumn{1}{c}{Length} & \multicolumn{1}{c}{Class} \\
    \midrule
    EC   & EthanolConcentration & 261  & 263  & 3    & 1751 & 4 \\
    FD   & FaceDetection & 5890 & 3524 & 144  & 62   & 2 \\
    HW   & Handwriting & 150  & 850  & 3    & 152  & 26 \\
    HB   & Heartbeat & 204  & 205  & 61   & 405  & 2 \\
    JV   & JapaneseVowels & 270  & 370  & 12   & 29   & 9 \\
    PS   & PEMS-SF & 267  & 173  & 963  & 144  & 7 \\
    SCP1 & SelfRegulationSCP1 & 268  & 293  & 6    & 896  & 2 \\
    SCP2 & SelfRegulationSCP2 & 200  & 180  & 7    & 1152 & 2 \\
    SAD  & SpokenArabicDigits & 6599 & 2199 & 13   & 93   & 10 \\
    UGL  & UWaveGestureLibrary & 120  & 320  & 3    & 315  & 8 \\
    \bottomrule
    \end{tabular}%

  \label{tab:datasets}%
\end{table*}%

\subsubsection{Baselines}
We included 11 state-of-the-art baselines from different categories for a comprehensive comparison. 

\begin{itemize}
    \item \textbf{MLP-based Method}
    \begin{itemize}
        \item DLinear~\cite{zeng2023transformers}: a combination of a decomposition scheme used in Autoformer and FEDformer with linear layers.
    \end{itemize}
    \item \textbf{CNN-based Method}
    \begin{itemize}
        \item TimesNet~\cite{wu2022timesnet}: uses a parameter-efficient inception block to discover multiple periods and capture temporal 2D-variations.
    \end{itemize}
    \item \textbf{Transformer-based Methods}
    \begin{itemize}
        \item Autoformer~\cite{wu2021autoformer}: a decomposition architecture with an Aoto-Correlation mechanism.
        \item Crossformer~\cite{zhang2023crossformer}: a Transformer-based model utilizing cross-dimension dependency for MTS forecasting.
        \item FEDformer~\cite{zhou2022fedformer}: a frequency enhanced decomposed transformer.
        \item Informer~\cite{zhou2021informer}: an efficient transformer-based model with a ProbSparse self-attention module.
        \item PatchTST~\cite{nie2022time}: uses segmentation of time series into subseries-level patches and channel-independence mechanism. 
        \item Reformer~\cite{kitaev2020reformer}: uses locality-sensitive hashing and reversible residual layers for efficient transformer.
        \item Vanilla Transformer~\cite{vaswani2017attention}: the classical transformer architecture.
    \end{itemize}
    \item \textbf{Contrastive-based Method}
    \begin{itemize}
        \item TF-C~\cite{zhang2022self}: a Time-Frequency Consistency mechanism for contrastive learning.
    \end{itemize}
    \item \textbf{LLM-based Method}
    \begin{itemize}
        \item OneFitsAll~\cite{zhou2023one}: a unified framework that uses a frozen pre-trained language model for all major types of time series analysis tasks.
    \end{itemize}
\end{itemize}

Some of the baselines were originally developed for forecasting tasks and later extended to classification tasks. To make a fair comparison, we implement baselines with their official codes\footnote{https://github.com/thuml/Time-Series-Library}\footnote{https://github.com/DAMO-DI-ML/NeurIPS2023-One-Fits-All}\footnote{https://github.com/mims-harvard/TFC-pretraining}. 
TF-C is designed for the pre-training and fine-tuning paradigm with few-shot scenarios.
We do not include more baselines designed for few-shot learning because most of them are for a specific field, making it difficult to apply to 10 extensive datasets.
The other baselines are classical deep learning models, so we modified the training data loader to randomly select a subset of training samples to simulate an N-way-K-shot classification problem. 

\subsubsection{Few-shot Evaluation}
We use classification accuracy to evaluate models and calculate an average accuracy to comprehensively demonstrate model capabilities. We conduct an N-way-K-shot classification task. In K-shot learning scenarios, we randomly select K samples from each class to conduct a K-shot training dataset. We report the accuracy performance only on the test dataset, which is the same as the classic evaluation for model training on the full train dataset. 

\subsubsection{Implementation Details}
We conduct all experiments on a Linux interactive high-performance computing machine with 2 Nvidia A40 GPUs, each with 48GB of memory and CUDA 11.8. We use bfloat16 to speed up the LoRA fine-tuning and inference of LLMs and reduce the memory usage simultaneously. We set the training epoch as 200 and the learning rate as 0.0002 with decaying 80\% every 50 epochs. All reported experimental results are averages of over 5 runs to avoid the impact of randomness brought by hardware equipment. See Section 3 in the Appendix for more details on implementation and hyperparameter settings.

\begin{table*}[t]
  \centering
   \setlength{\tabcolsep}{0.5mm}
   \caption{Comparison of classification accuracy (\%) under 1-shot setting. All performances are the average over 5 runs. \textit{Improv. \%} is the performance improvement of our model over the second-best baseline. * denotes a significant improvement over the best baseline ($P <0.05$). \textbf{Boldface}: Best results; \underline{Underlined}: Second-best results.}
   {\small
    \begin{tabular}{lrrrlrrlrrr}
    \toprule
         Accuracy $\uparrow$& \multicolumn{1}{c}{EC} & \multicolumn{1}{c}{FD} & \multicolumn{1}{c}{HW} & \multicolumn{1}{c}{HB} & \multicolumn{1}{c}{JV} & \multicolumn{1}{c}{PS} & \multicolumn{1}{c}{SCP1} & \multicolumn{1}{c}{SCP2} & \multicolumn{1}{c}{SAD} & \multicolumn{1}{c}{UGL} \\
    \midrule
    DLinear & {$28.0_{\pm 0.2}$} & \underline {$51.4_{\pm 1.8}$} & {$10.3_{\pm 3.1}$} & \underline{$73.0_{\pm 1.6}$} & {$61.8_{\pm 1.1}$} & {$28.9_{\pm 1.4}$} & {$87.4_{\pm 0.4}$} & {$51.3_{\pm 4.5}$} & {$29.3_{\pm 4.2}$} & {$59.9_{\pm 4.1}$} \\
    TimesNet & \underline {$28.4_{\pm 1.0}$} & {$50.1_{\pm 0.4}$} & {$11.8_{\pm 0.5}$} & {$58.2_{\pm 14.7}$} & {$52.3_{\pm 3.3}$} & {$27.2_{\pm 4.5}$} & {$86.3_{\pm 1.5}$} & {$50.4_{\pm 3.1}$} & {$30.4_{\pm 4.3}$} & {$57.4_{\pm 7.7}$} \\
    Autoformer & {$27.0_{\pm 1.0}$} & {$50.6_{\pm 1.1}$} & {$8.2_{\pm 2.2}$} & {$48.7_{\pm 4.0}$} & {$38.7_{\pm 5.5}$} & {$30.2_{\pm 2.5}$} & {$52.4_{\pm 0.9}$} & {$53.0_{\pm 2.2}$} & {$39.2_{\pm 3.8}$} & {$23.8_{\pm 5.3}$} \\
    Crossformer & {$27.5_{\pm 0.7}$} & {$51.2_{\pm 1.0}$} & {$11.2_{\pm 4.0}$} & {$72.1_{\pm 1.8}$} & \underline {$70.7_{\pm 0.8}$} & {$35.4_{\pm 7.8}$} & {$87.7_{\pm 1.1}$} &  {$54.3_{\pm 4.3}$} & \underline{$53.3_{\pm 7.9}$} & {$58.6_{\pm 5.8}$} \\
    FEDformer & {$27.0_{\pm 0.8}$} & {$50.5_{\pm 0.7}$} & {$9.6_{\pm 1.3}$} & {$72.7_{\pm 1.0}$} & {$49.6_{\pm 4.1}$} & {$24.2_{\pm 7.6}$} & {$50.9_{\pm 1.6}$} & {$53.8_{\pm 0.5}$} & {$37.5_{\pm 5.0}$} & {$27.6_{\pm 1.7}$} \\
    Informer & {$26.1_{\pm 0.6}$} & {$50.3_{\pm 0.4}$} & {$13.0_{\pm 1.3}$} & {$69.3_{\pm 6.5}$} & {$64.6_{\pm 1.6}$} & {$31.3_{\pm 1.7}$} & {$87.7_{\pm 0.0}$} & {$47.7_{\pm 0.6}$} & {$51.3_{\pm 2.0}$} & {$60.4_{\pm 5.9}$} \\
    PatchTST & {$27.1_{\pm 2.0}$} & {$50.7_{\pm 0.6}$} & {$12.9_{\pm 0.8}$} & {$66.3_{\pm 4.3}$} & {$57.6_{\pm 5.5}$} & \underline {$41.0_{\pm 2.3}$} & {$58.4_{\pm 3.7}$} & {$53.1_{\pm 1.1}$} & {$45.7_{\pm 4.0}$} & {$58.4_{\pm 6.1}$} \\
    Reformer & {$28.3_{\pm 2.1}$} & {$50.0_{\pm 0.2}$} & {$13.5_{\pm 0.3}$} & {$52.8_{\pm 6.1}$} & {$61.8_{\pm 8.1}$} & {$31.9_{\pm 3.0}$} & \underline{$87.9_{\pm 0.7}$} & {$51.4_{\pm 2.1}$} & {$51.7_{\pm 4.0}$} & {$58.8_{\pm 4.5}$} \\
    Transformer & {$26.5_{\pm 0.3}$} & {$50.7_{\pm 0.5}$} & \underline {$15.2_{\pm 0.8}$} & {$61.1_{\pm 13.6}$} & {$63.6_{\pm 8.6}$} & {$31.2_{\pm 2.9}$} & {$69.3_{\pm 19.1}$} & {$46.8_{\pm 0.5}$} & {$49.8_{\pm 4.9}$} & {$57.4_{\pm 5.5}$} \\
    TF-C & $32.5_{\pm 0.9}$ & $51.0_{\pm 0.1}$ & $9.5_{\pm 0.4}$ & $72.8_{\pm 0.2}$ & $21.6_{\pm 2.2}$ & $26.5_{\pm 2.6}$ & $53.0_{\pm 0.6}$ & \underline{$56.0_{\pm 0.6}$} & $15.4_{\pm 1.6}$ & $29.0_{\pm 3.0}$ \\
    OneFitsAll & {$28.1_{\pm 1.4}$} & {$50.9_{\pm 0.5}$} & {$13.4_{\pm 0.8}$} & {$63.9_{\pm 14.0}$} & {$66.3_{\pm 5.2}$} & {$35.7_{\pm 1.4}$} & {$62.6_{\pm 13.8}$} & {$52.9_{\pm 2.5}$} & {$44.2_{\pm 5.6}$} & \underline {$64.7_{\pm 3.6}$} \\
    LLMFew & $\mathbf{42.6_{\pm 1.9}}$ & $\mathbf{66.9_{\pm 6.6}}$ & $\mathbf{34.3_{\pm 0.9}}$ & $\mathbf{79.8_{\pm 6.5}}$ & $\mathbf{78.6_{\pm 2.0}}$ & $\mathbf{45.9_{\pm 4.6}}$ & $\mathbf{88.5_{\pm 1.9}}$ & $\mathbf{58.1_{\pm 2.5}}$ & $\mathbf{61.4_{\pm 1.9}}$ & $\mathbf{76.2_{\pm 1.8}}$ \\
    \midrule
    \textit{Improv. \% } &\textit{ 50.2\%*} & \textit{30.3\%*} & \textit{125.2\%*} & \multicolumn{1}{r}{\textit{9.2\%*}} & \textit{11.3\%*} & \textit{11.8\%*} & \multicolumn{1}{r}{\textit{0.7\%}} & \textit{3.8\%} & \textit{15.1\%*} & \textit{17.8\%*} \\
    \bottomrule
    \end{tabular}%
    }
  \label{tab:main}%
\end{table*}%

\subsection{Result Analysis}
\subsubsection{Overall Performance}
Table~\ref{tab:main} compares the classification accuracy of LLMFew against all baselines on 10 UEA real-world datasets under the 1-shot learning setting, i.e. only 1 sample per class in the training dataset. To make a fair comparison with OneFitsAll, in Table~\ref{fig:framework}, the results for LLMFew are implemented by GPT2-117M, not the best performance among all kinds of LLMs. We discuss the performance difference in the latter Section. Overall, LLMFew outperforms all other baselines. Notably, LLMFew is particularly effective on challenging datasets, especially for HW and EC datasets, with 152.2\% and 50.2\% improvements, respectively. These two datasets can be considered hard because HW has the most classes of 26, and EC includes an extremely long time series at the length of 1751, compared to the other datasets in the benchmark. The satisfactory performance with EC also shows that our model can activate LLM's long sequence modeling ability in the time series field. However, LLMFew has not achieved a significant improvement on the PS dataset. PS is a traffic dataset characterized by exceptionally high dimensionality with 963 features. This phenomenon prompts us to consider how to model the relationships between features in the MTSC task as dimensions increase, particularly to support classification outcomes in domains like traffic datasets where the relationships between dimensions are very closely intertwined. 

Additionally, there are some important insights related to the baselines. The well-known end-to-end Transformer-based time series models have failed to achieve satisfactory performance, demonstrating that pre-training knowledge in LLMs is essential and that the Transformer structure alone is insufficient for few-shot MTSC tasks. 
Besides, the TF-C model, initially pre-trained on a sleep ECG dataset, was fine-tuned with our dataset. Despite this, the results were highly variable across different datasets. It delivered a satisfactory performance on SCP2, ranking second, but struggled significantly with the HW dataset, showing almost no learning. This observation underscores a critical challenge in few-shot learning with traditional approaches—the distributional disparity between the source and target datasets. 
Another noteworthy observation is that OneFitsAll, which utilizes a partially frozen GPT as its backbone, does not exhibit outstanding performance in few-shot scenarios. In this approach, only the positional embedding and normalization layers are trained, making the operation more akin to directly mapping time series representations to the text semantic space of LLMs.
This direct design is efficient but lacks interpretability. LLMs struggle to understand modalities that are not present in their pre-trained datasets. This highlights the necessity of fine-tuning, as LLMs must learn unique information from time series data to effectively integrate it with their existing world knowledge.

\subsubsection{K-shot Learning Performance}

\begin{figure}[t]
\centering
\includegraphics[width=0.75\columnwidth]{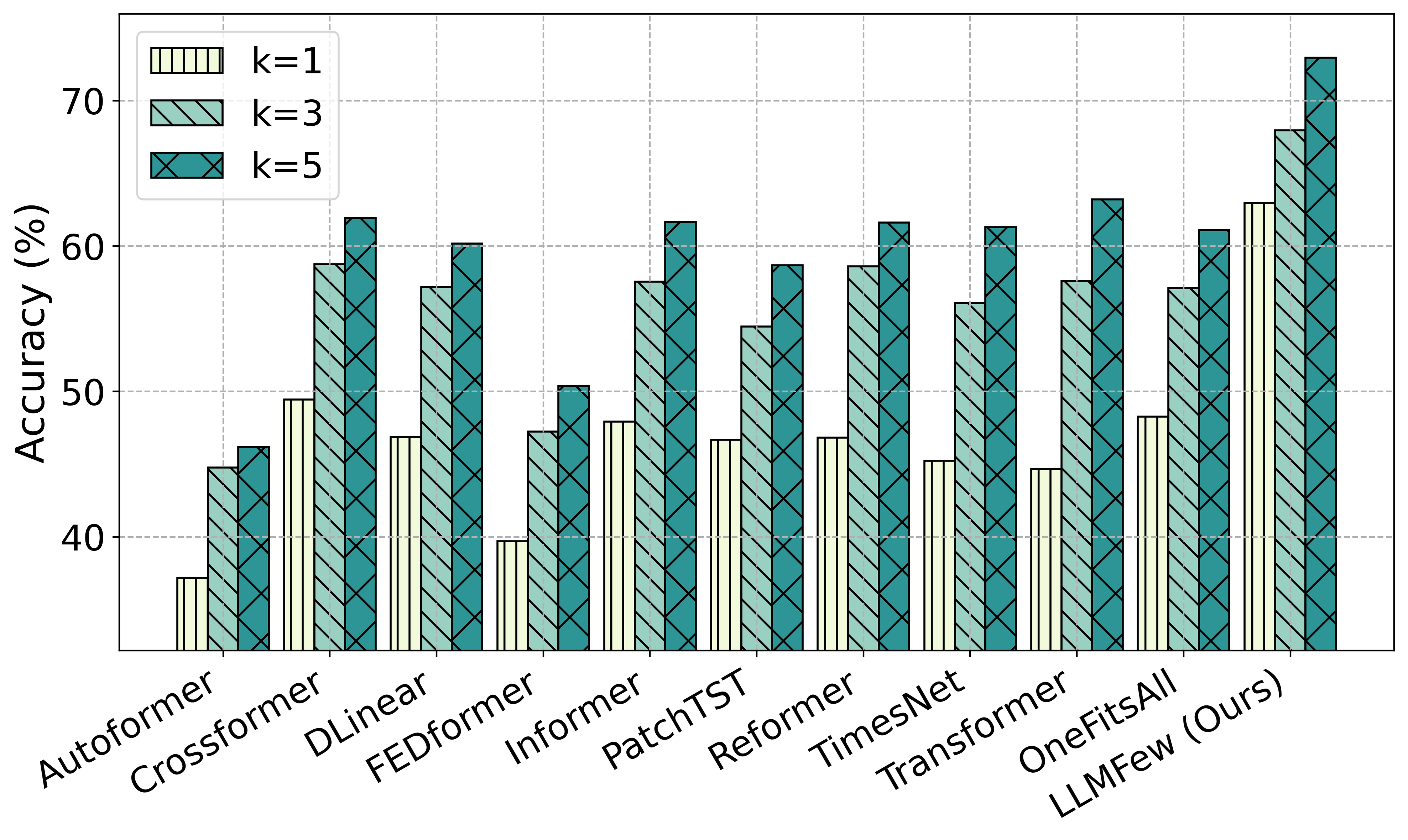} 
\caption{Average classification accuracy (\%) under K-shot setting, with $k=1,3,5$. All results are the average over 5 runs.}

\label{fig:nshot}
\end{figure}

Figure~\ref{fig:nshot} shows the average classification accuracy for baselines and LLMFew with K-shot setting\footnote{In the HW dataset, some classes have only 2 samples in the training set. For the 3-shot and 5-shot settings, we include all available samples from these classes (i.e., possibly just 2) to generate the training dataset.}.
It can be concluded that, for all models, performance improves as k increases. However, an interesting observation is that in the few-shot learning scenario, the marginal benefit of adding more samples diminishes.
As an example, LLMFew attains an average accuracy of 62.95\% in the 1-shot setting, which improves to 72.96\% in the 5-shot setting. However, even with the full training dataset, the average accuracy only rises to 78.38\%.
A similar conclusion can be drawn for other baselines when comparing their 1-shot, 3-shot, and 5-shot performances. However, their 5-shot performance still lags behind the results obtained using the full training dataset. One possible explanation is that end-to-end deep models trained from scratch require more samples to learn data features and generalize to unseen test samples. In contrast, our pipeline fine-tunes only a small number of parameters, allowing the LLM to retain and transfer its ability to model sequence features, achieving satisfactory accuracy without needing more samples. 

\subsubsection{Results with Full Train Datasets}

\begin{figure}[t]
\centering
\includegraphics[width=0.75\columnwidth]{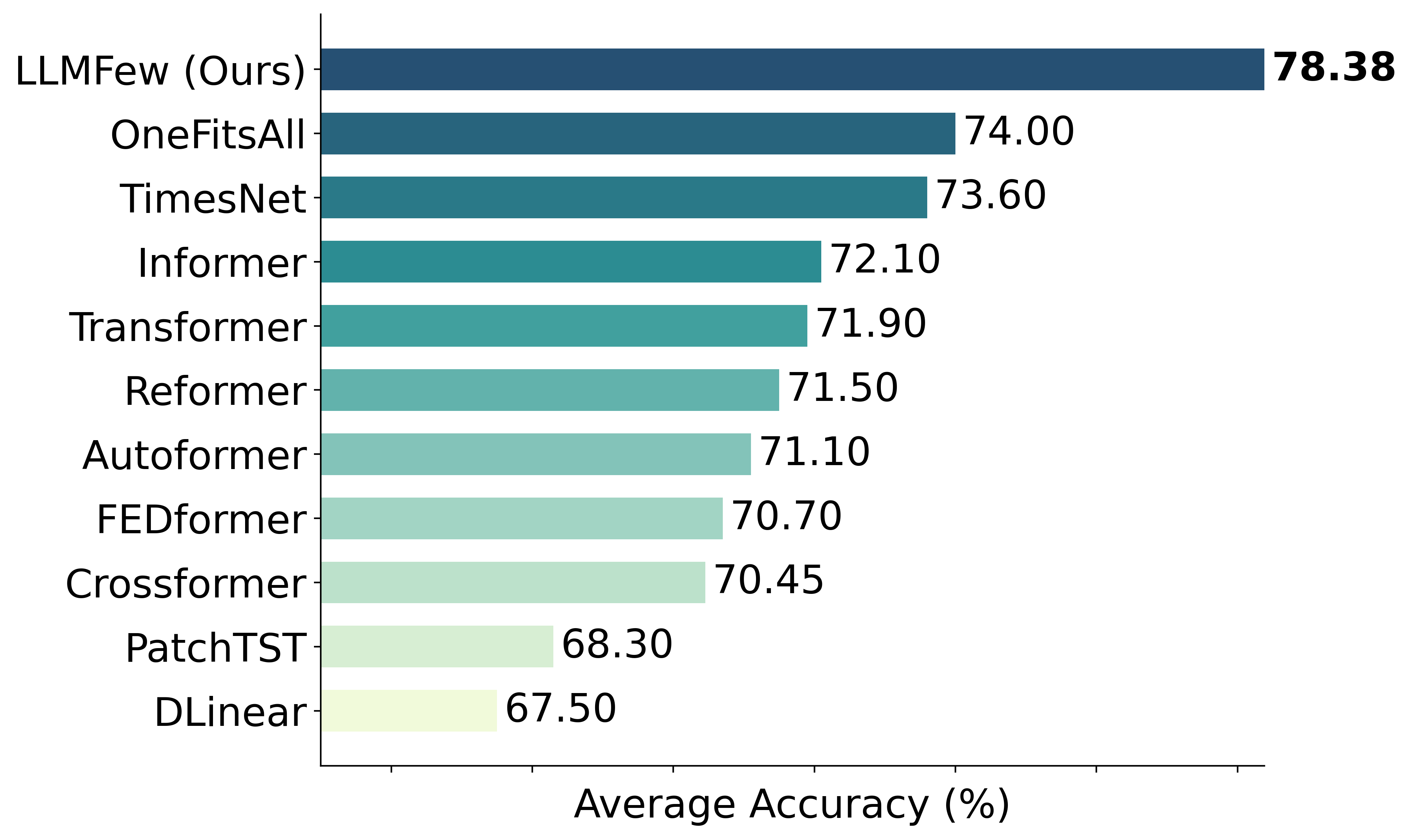} 
\caption{Average classification accuracy (\%) with full training samples. The results for LLMFew, Crossformer and PatchTST are the average over 5 runs. The others are reported from OneFitsAll.}
\label{fig:fulltrain}
\end{figure}

Figure~\ref{fig:fulltrain} presents the results using the classical training paradigm, where the model is trained on all available samples and evaluated on the test dataset. This is the same evaluation method used for classical end-to-end models in~\cite{zhou2023one}.
This analysis was conducted to evaluate how our model’s performance evolves as the number of training samples increases (beyond just a few) and to determine if it can effectively learn from these additional samples. The model is expected to demonstrate robustness to varying sample sizes. LLMFew consistently outperforms all state-of-the-art baselines, proving that its performance is both satisfactory and reliably stable for MTSC tasks.
Figure~\ref{fig:fulltrain} displays the average classification accuracy for each model across 10 datasets, representing the overall performance of the models. 

\begin{table*}[t]
  \centering
  \setlength{\tabcolsep}{1.3mm} 
  \caption{Comparison of classification accuracy (\%) for LLMFew variants with different size LLMs under the 1-shot setting. All results are the average over 5 runs. \textbf{Boldface}: Best results; \underline{Underlined}: Second-best results.}  
    \begin{tabular}{llllllll}
    \toprule
         & \textit{GPT2{\scriptsize-s}} & \textit{GPT2{\scriptsize-m}} & \textit{GPT2{\scriptsize-l}} & \textit{Qwen{\scriptsize-0.5B}} & \textit{Qwen{\scriptsize-1.5B}} & \textit{Phi3} & \textit{Llama3} \\
    \midrule
    EC & $42.6_{\pm 1.9}$ & $\underline{46.1_{\pm 5.5}}$ & $\mathbf{52.1_{\pm 7.4}}$ & $39.8_{\pm 6.2}$ & $40.7_{\pm 10.4}$ & $37.9_{\pm 6.6}$ & $39.1_{\pm 9.1}$ \\
    FD & $\underline{66.9_{\pm 6.6}}$ & $60.2_{\pm 8.9}$ & $57.2_{\pm 4.3}$ & $54.6_{\pm 4.3}$ & $52.5_{\pm 1.1}$ & $65.1_{\pm 8.5}$ & $\mathbf{68.6_{\pm 7.9}}$ \\
    HW & $34.3_{\pm 0.9}$ & $34.4_{\pm 2.1}$ & $\underline{36.3_{\pm 3.3}}$ & $31.5_{\pm 3.4}$ & $\mathbf{36.4_{\pm 4.9}}$ & $34.3_{\pm 3.3}$ & $28.5_{\pm 1.8}$ \\
    HB & $\mathbf{79.8_{\pm 6.5}}$ & $67.6_{\pm 1.3}$ & $\underline{72.9_{\pm 1.3}}$ & $61.2_{\pm 5.7}$ & $70.6_{\pm 4.5}$ & $69.2_{\pm 8.7}$ & $65.3_{\pm 1.1}$ \\
    JV & $\underline{78.6_{\pm 2.0}}$ & $\mathbf{78.7_{\pm 2.4}}$ & $74.2_{\pm 2.5}$ & $69.7_{\pm 4.6}$ & $77.3_{\pm 3.5}$ & $76.0_{\pm 2.4}$ & $78.3_{\pm 2.7}$ \\
    PS & $45.9_{\pm 4.6}$ & $45.0_{\pm 4.6}$ & $47.5_{\pm 5.4}$ & $40.7_{\pm 7.4}$ & $40.1_{\pm 8.1}$ & $\mathbf{49.6_{\pm 6.9}}$ & $\underline{48.8_{\pm 6.4}}$ \\
    SCP1 & $\mathbf{88.5_{\pm 1.9}}$ & $84.7_{\pm 4.4}$ & $86.2_{\pm 3.7}$ & $82.3_{\pm 2.8}$ & $85.8_{\pm 4.1}$ & $83.0_{\pm 1.6}$ & $\underline{86.4_{\pm 3.2}}$ \\
    SCP2 & $58.1_{\pm 2.5}$ & $60.4_{\pm 3.4}$ & $60.9_{\pm 0.6}$ & $52.3_{\pm 6.2}$ & $54.1_{\pm 11.7}$ & $\underline{64.0_{\pm 8.0}}$ & $\mathbf{75.2_{\pm 0.1}}$ \\
    SAD & $61.4_{\pm 1.9}$ & $\underline{66.0_{\pm 8.4}}$ & $64.3_{\pm 6.1}$ & $53.0_{\pm 8.0}$ & $54.9_{\pm 4.7}$ & $63.1_{\pm 2.2}$ & $\mathbf{66.7_{\pm 5.7}}$ \\
    UGL & $76.2_{\pm 1.8}$ & $72.4_{\pm 5.4}$ & $69.8_{\pm 8.9}$ & $74.7_{\pm 4.0}$ & $73.4_{\pm 5.1}$ & $\mathbf{78.2_{\pm 2.3}}$ & $\underline{76.9_{\pm 3.3}}$ \\
    \bottomrule
    \end{tabular}%
  \label{tab:llms}%
\end{table*}%

\subsubsection{Performance of Different Sizes of LLMs}

\label{sec:diffllms}

Table~\ref{tab:llms} shows the classification accuracy of LLMFew variants with different LLMs, including GPT2 small, medium, and large versions (117M, 345M, and 774M)~\cite{radford2019language}, Qwen (0.5B and 1.5B)~\cite{bai2023qwen}, Phi3 (3.8B)~\cite{abdin2024phi} and Llama3 (8B)~\cite{dubey2024llama}. 
To ensure a fair comparison among the LLM variants, we used LoRA to fine-tune the same set of parameters, specifically the Q, K, and V parameters within the attention layers. We selected LLMs with varying parameter sizes to explore the potential for a scaling law~\cite{biderman2023pythia}, where LLMs with more parameters might demonstrate superior performance on our specific topic.
Because of GPU memory limitations, we did not include larger LLMs (e.g., Llama2-13B) in our analysis. Across all datasets, it's difficult to draw a straightforward conclusion regarding the relationship between LLM size and performance. However, the choice of LLM might depend on the specific dataset. For example, the JV dataset, which has a dimension of 12 and a length of 29, achieves state-of-the-art performance with an LLM of approximately 300M parameters. Conversely, the PS dataset, with a dimension of 963 and a length of 144, requires larger LLMs to achieve better accuracy.


\begin{table*}[!t]
  \centering
  \caption{Ablation Study of classification accuracy (\%) for all datasets. All results are the average over 5 runs. \textbf{Boldface}: Best results; \underline{Underlined}: Second-best results.}
    \begin{tabular}{lllllllll}
    \toprule
    & \textit{LLMFew} & \textit{w/o PTCEnc} & \textit{Frozen} & \textit{w/o LLM} \\
    \midrule
    EC & $\mathbf{42.6_{\pm 1.9}}$ & $33.7_{\pm 4.1}$ & \underline{$36.7_{\pm 3.8}$} & $32.9_{\pm 8.5}$ \\
    FD & $\mathbf{66.9_{\pm 6.6}}$ & $55.7_{\pm 3.2}$ & \underline{$55.9_{\pm 3.1}$} & $52.7_{\pm 3.3}$ \\
    HW & $\mathbf{34.3_{\pm 0.9}}$ & $19.3_{\pm 2.2}$ & \underline{$31.6_{\pm 3.6}$} & $30.6_{\pm 3.7}$ \\
    HB & $\mathbf{79.8_{\pm 6.5}}$ & $46.6_{\pm 12.9}$ & $51.1_{\pm 9.6}$ & \underline{$58.0_{\pm 12.4}$} \\
    JV & $\mathbf{78.6_{\pm 2.0}}$ & $72.7_{\pm 4.3}$ & $74.0_{\pm 4.6}$ & \underline{$75.1_{\pm 0.0}$} \\
    PS & $\mathbf{45.9_{\pm 4.6}}$ & \underline{$37.0_{\pm 7.3}$} & $36.8_{\pm 4.0}$ & $35.3_{\pm 5.9}$ \\
    SCP1 & $\mathbf{88.5_{\pm 1.9}}$ & \underline{$83.3_{\pm 1.2}$} & $80.5_{\pm 5.4}$ & $84.3_{\pm 4.8}$ \\
    SCP2 & $\mathbf{58.1_{\pm 2.5}}$ & $49.2_{\pm 4.5}$ & \underline{$54.7_{\pm 7.6}$} & $54.2_{\pm 5.5}$ \\
    SAD & $\mathbf{61.4_{\pm 1.9}}$ & \underline{$60.6_{\pm 2.7}$} & $59.0_{\pm 2.1}$ & $59.6_{\pm 2.9}$ \\
    UGL & $\mathbf{76.2_{\pm 1.8}}$ & $68.2_{\pm 7.1}$ & \underline{$73.2_{\pm 2.2}$} & $70.6_{\pm 2.9}$ \\
    \bottomrule
    \end{tabular}%
  \label{tab:ablation}%
\end{table*}

\subsubsection{Ablation Study}
To evaluate the effectiveness of each component and its contribution to overall classification accuracy, we developed three variants of LLMFew, with their accuracy reported in Table~\ref{tab:ablation}.
The \textit{w/o PTCEnc} variant replaces the PTCEnc with a simple 1D convolution module, aligning the input feature dimensions with the LLM embedding dimensions. The \textit{Frozen} variant bypasses LoRA for fine-tuning the LLM's parameters by freezing them and instead trains only the PTCEnc and classification head. Lastly, the \textit{w/o LLM} variant removes the LLM decoder, feeding the time series embedding (i.e., the output of the PTCEnc) directly to the classification head.
With the removal of each component from our framework, performance degrades to varying degrees. Across all four datasets, \textit{Frozen} achieves the second-best performance, demonstrating that the LLM's pre-trained capabilities are valuable in MTSC, even without fine-tuning its parameters. When comparing with \textit{w/o LLM}, it's evident that the frozen LLM decoder plays a significant role in enhancing overall performance. Additionally, the PTCEnc plays a crucial role in effectively learning and aligning the MTS input with the LLM's embedding space, particularly in the HW dataset with the largest number of classes.

\subsubsection{Inference Time and Memory Cost}

\begin{figure}[t]
\centering
\includegraphics[width=0.75\columnwidth]{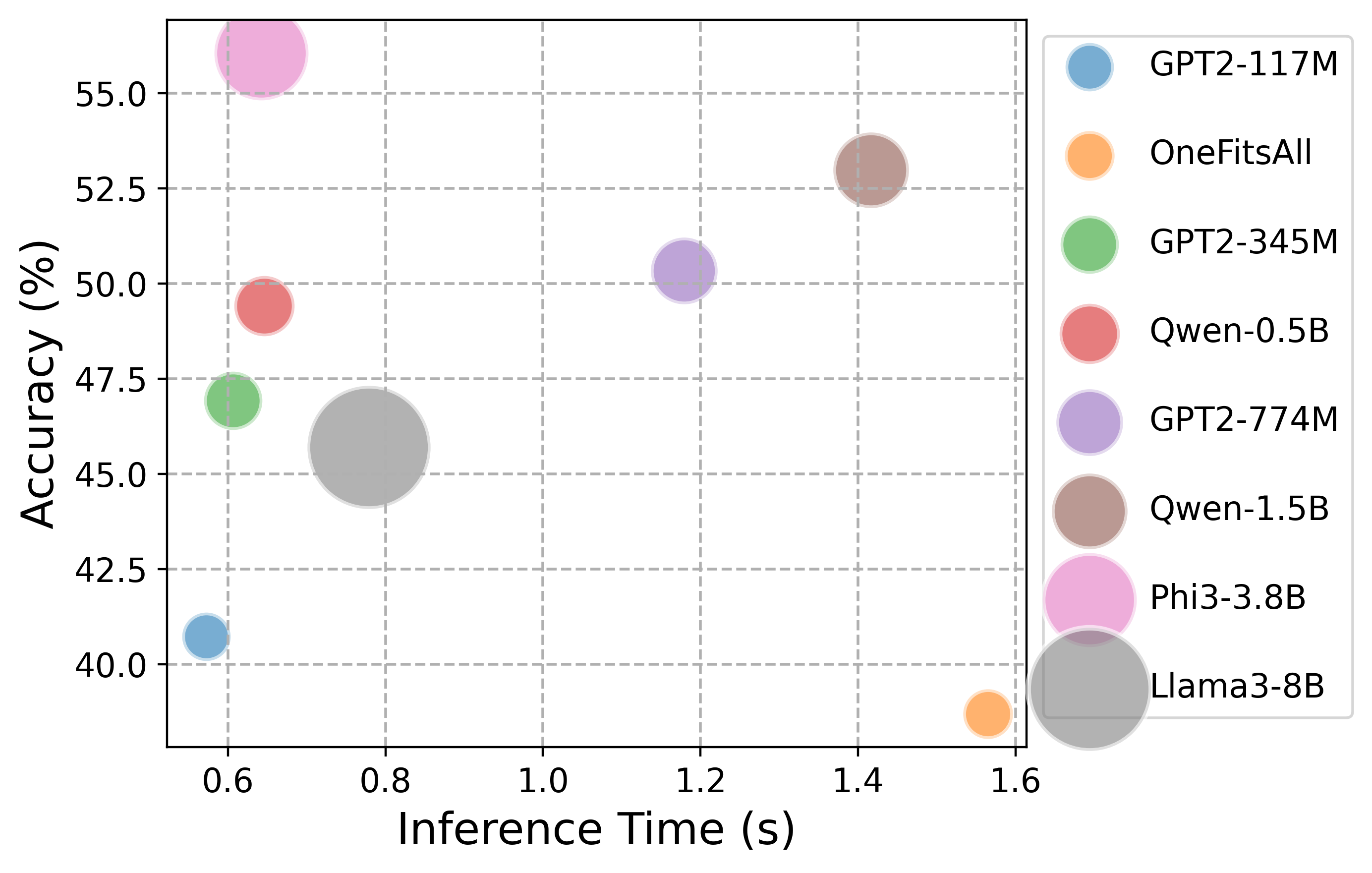} 
\caption{Inference time and GPU memory usage comparison for LLM-based methods. The results are for the PEMS-SF dataset. OneFitsAll is one of our baselines. The others are LLMFew variants, same as them in Table~\ref{tab:llms} The results are average over 5 runs.}
\label{fig:memory}
\end{figure}

Figure~\ref{fig:memory} shows the inference time and memory usage for LLM-based methods (OneFitsAll and LLMFew variants) on the PEMS-SF dataset, which can be considered as the most difficult one among 10 datasets. We also report the average classification accuracy for all models. The satisfactory model should have competitive performance with low memory usage and short inference time. So, we expect the model to appear in the upper left corner with a small dot. 
It is worth noting that on the PEMS-SF dataset, it is not the case that the larger the model, the better the performance.
This is inconsistent with the intuitive scaling law for LLMs~\cite{biderman2023pythia}. In Figure 5, the performance of the phi3 3.8b variant even exceeds that of the llama3 8b, which has more than twice as many parameters. One possible speculation is that the excellent performance of phi3 in the inter-GPS test~\cite{lu2021inter} helps it understand the geospatial relationship among features in the PEMS-SF dataset.

\subsubsection{Hyperparameter Sensitivity Analysis} 

\begin{figure}[t]
\centering
\includegraphics[width=0.95\columnwidth]{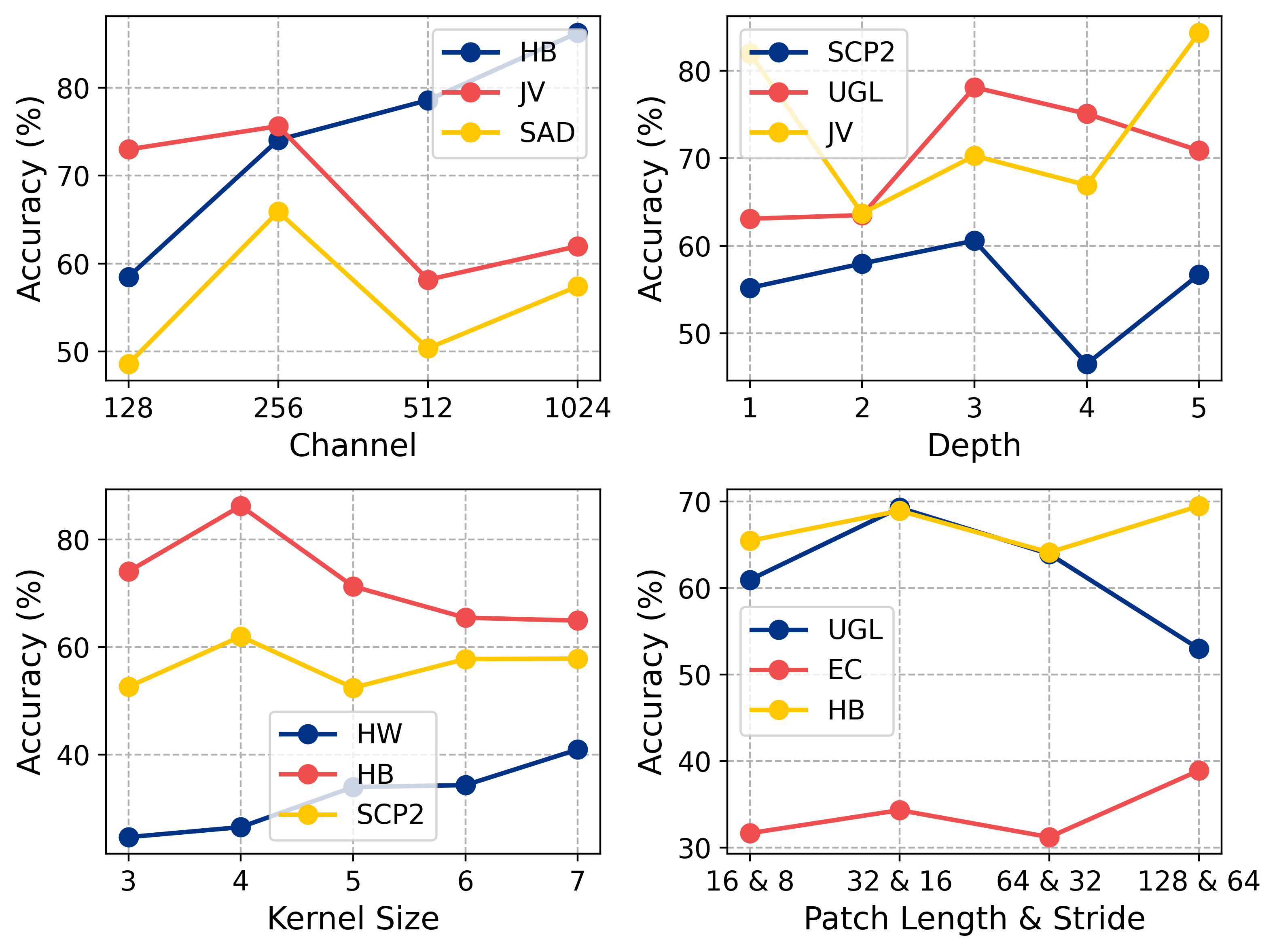} 
\caption{Key hyperparameter sensitivity analysis for the patching layer and PTCEnc on selected datasets.}
\label{fig:hyper}
\end{figure}

This section aims to explore how specific hyperparameters in the patching layer and PTCEnc influence the model’s performance.
Figure~\ref{fig:hyper} illustrates the impact of key hyperparameters in the patching layer and PTCEnc. We choose 4 different values for the parameters and evaluate their performance. We choose $\{128,256,512,1024\}$ for the convolution hidden state size, $\{1,2,3,4,5\}$ for the  depth of stacked layers, $\{3,4,5,6,7\}$ for the convolution kernel size, and $\{(16,8), (32,16),(64,32),(128,64)\}$ for the patch length and stride.
Different datasets exhibit varying optimal hidden channel sizes depending on their sizes and complexities. Larger hidden channel sizes can capture more information but may lead to overfitting and slower training speeds.
A similar conclusion applies to the depth of the temporal convolution structure. Deeper layers have larger dilation factors and wider receptive fields, enabling them to learn coarse and global temporal relationships. However, as the number of layers increases, the model also becomes more prone to overfitting.
The kernel size directly influences the size of the receptive field. A kernel size that is too large may overlook local features and, due to the increased number of convolution parameters, can easily lead to overfitting, especially on datasets with limited training samples, such as in few-shot scenarios. In our experiments, a kernel size of 3 proved to be the best choice for most datasets.
When selecting patch length and stride, we opted not to conduct a grid search and instead set the stride to half the patch length. While longer patches capture more long-term dependencies in the time dimension, the periodicity of this dimension depends on factors like data acquisition methods. As a result, we have yet to find a patch length that is universally applicable to all datasets.

\section{Conclusion}
In this work, we address an overlooked yet critical gap in Multivariate Time Series Classification (MTSC): the few-shot learning scenarios, which are especially important when training data is limited, a common scenario in many industrial environments where obtaining labeled data is costly and time-consuming. To bridge this gap, we introduce LLMFew, a model designed to explore the potential of Large Language Models (LLMs) in few-shot MTSC tasks. LLMFew leverages the powerful pre-trained representations of LLMs, enabling effective generalization from minimal data and overcoming the bottleneck faced by traditional models in low-data regimes. Our experiments on real-world datasets—including those for gesture, action, and audio recognition—demonstrate that LLMFew effectively addresses the data scarcity problem in few-shot learning. The insights gained from our study offer a promising direction for applying LLMs to other few-shot time series tasks. Moving forward, we aim to enhance the interpretability of LLMs in time series applications and further investigate which aspects of LLMs are most effective for learning temporal representations in multivariate time series.

\bibliography{sn-bibliography}


\begin{thebibliography}{43}
\ifx \bisbn   \undefined \def \bisbn  #1{ISBN #1}\fi
\ifx \binits  \undefined \def \binits#1{#1}\fi
\ifx \bauthor  \undefined \def \bauthor#1{#1}\fi
\ifx \batitle  \undefined \def \batitle#1{#1}\fi
\ifx \bjtitle  \undefined \def \bjtitle#1{#1}\fi
\ifx \bvolume  \undefined \def \bvolume#1{\textbf{#1}}\fi
\ifx \byear  \undefined \def \byear#1{#1}\fi
\ifx \bissue  \undefined \def \bissue#1{#1}\fi
\ifx \bfpage  \undefined \def \bfpage#1{#1}\fi
\ifx \blpage  \undefined \def \blpage #1{#1}\fi
\ifx \burl  \undefined \def \burl#1{\textsf{#1}}\fi
\ifx \doiurl  \undefined \def \doiurl#1{\url{https://doi.org/#1}}\fi
\ifx \betal  \undefined \def \betal{\textit{et al.}}\fi
\ifx \binstitute  \undefined \def \binstitute#1{#1}\fi
\ifx \binstitutionaled  \undefined \def \binstitutionaled#1{#1}\fi
\ifx \bctitle  \undefined \def \bctitle#1{#1}\fi
\ifx \beditor  \undefined \def \beditor#1{#1}\fi
\ifx \bpublisher  \undefined \def \bpublisher#1{#1}\fi
\ifx \bbtitle  \undefined \def \bbtitle#1{#1}\fi
\ifx \bedition  \undefined \def \bedition#1{#1}\fi
\ifx \bseriesno  \undefined \def \bseriesno#1{#1}\fi
\ifx \blocation  \undefined \def \blocation#1{#1}\fi
\ifx \bsertitle  \undefined \def \bsertitle#1{#1}\fi
\ifx \bsnm \undefined \def \bsnm#1{#1}\fi
\ifx \bsuffix \undefined \def \bsuffix#1{#1}\fi
\ifx \bparticle \undefined \def \bparticle#1{#1}\fi
\ifx \barticle \undefined \def \barticle#1{#1}\fi
\bibcommenthead
\ifx \bconfdate \undefined \def \bconfdate #1{#1}\fi
\ifx \botherref \undefined \def \botherref #1{#1}\fi
\ifx \url \undefined \def \url#1{\textsf{#1}}\fi
\ifx \bchapter \undefined \def \bchapter#1{#1}\fi
\ifx \bbook \undefined \def \bbook#1{#1}\fi
\ifx \bcomment \undefined \def \bcomment#1{#1}\fi
\ifx \oauthor \undefined \def \oauthor#1{#1}\fi
\ifx \citeauthoryear \undefined \def \citeauthoryear#1{#1}\fi
\ifx \endbibitem  \undefined \def \endbibitem {}\fi
\ifx \bconflocation  \undefined \def \bconflocation#1{#1}\fi
\ifx \arxivurl  \undefined \def \arxivurl#1{\textsf{#1}}\fi
\csname PreBibitemsHook\endcsname

\bibitem[\protect\citeauthoryear{Ismail~Fawaz et~al.}{2019}]{IsmailFawaz2018deep}
\begin{barticle}
\bauthor{\bsnm{Ismail~Fawaz}, \binits{H.}},
\bauthor{\bsnm{Forestier}, \binits{G.}},
\bauthor{\bsnm{Weber}, \binits{J.}},
\bauthor{\bsnm{Idoumghar}, \binits{L.}},
\bauthor{\bsnm{Muller}, \binits{P.-A.}}:
\batitle{Deep learning for time series classification: a review}.
\bjtitle{Data Mining and Knowledge Discovery}
\bvolume{33}(\bissue{4}),
\bfpage{917}--\blpage{963}
(\byear{2019})
\end{barticle}
\endbibitem

\bibitem[\protect\citeauthoryear{Rajkomar et~al.}{2018}]{rajkomar2018scalable}
\begin{barticle}
\bauthor{\bsnm{Rajkomar}, \binits{A.}},
\bauthor{\bsnm{Oren}, \binits{E.}},
\bauthor{\bsnm{Chen}, \binits{K.}},
\bauthor{\bsnm{Dai}, \binits{A.M.}},
\bauthor{\bsnm{Hajaj}, \binits{N.}},
\bauthor{\bsnm{Hardt}, \binits{M.}},
\bauthor{\bsnm{Liu}, \binits{P.J.}},
\bauthor{\bsnm{Liu}, \binits{X.}},
\bauthor{\bsnm{Marcus}, \binits{J.}},
\bauthor{\bsnm{Sun}, \binits{M.}}, \betal:
\batitle{Scalable and accurate deep learning with electronic health records}.
\bjtitle{NPJ digital medicine}
\bvolume{1}(\bissue{1}),
\bfpage{1}--\blpage{10}
(\byear{2018})
\end{barticle}
\endbibitem

\bibitem[\protect\citeauthoryear{Jiang et~al.}{2020}]{jiang2020decentralized}
\begin{barticle}
\bauthor{\bsnm{Jiang}, \binits{J.}},
\bauthor{\bsnm{Ji}, \binits{S.}},
\bauthor{\bsnm{Long}, \binits{G.}}:
\batitle{Decentralized knowledge acquisition for mobile internet applications}.
\bjtitle{World Wide Web}
\bvolume{23}(\bissue{5}),
\bfpage{2653}--\blpage{2669}
(\byear{2020})
\end{barticle}
\endbibitem

\bibitem[\protect\citeauthoryear{Fawaz et~al.}{2018}]{fawaz2018transfer}
\begin{bchapter}
\bauthor{\bsnm{Fawaz}, \binits{H.I.}},
\bauthor{\bsnm{Forestier}, \binits{G.}},
\bauthor{\bsnm{Weber}, \binits{J.}},
\bauthor{\bsnm{Idoumghar}, \binits{L.}},
\bauthor{\bsnm{Muller}, \binits{P.-A.}}:
\bctitle{Transfer learning for time series classification}.
In: \bbtitle{2018 IEEE International Conference on Big Data (Big Data)},
pp. \bfpage{1367}--\blpage{1376}
(\byear{2018}).
\bcomment{IEEE}
\end{bchapter}
\endbibitem

\bibitem[\protect\citeauthoryear{Zhang et~al.}{2022}]{zhang2022self}
\begin{barticle}
\bauthor{\bsnm{Zhang}, \binits{X.}},
\bauthor{\bsnm{Zhao}, \binits{Z.}},
\bauthor{\bsnm{Tsiligkaridis}, \binits{T.}},
\bauthor{\bsnm{Zitnik}, \binits{M.}}:
\batitle{Self-supervised contrastive pre-training for time series via time-frequency consistency}.
\bjtitle{Advances in Neural Information Processing Systems}
\bvolume{35},
\bfpage{3988}--\blpage{4003}
(\byear{2022})
\end{barticle}
\endbibitem

\bibitem[\protect\citeauthoryear{Tang et~al.}{2020}]{tang2020interpretable}
\begin{bchapter}
\bauthor{\bsnm{Tang}, \binits{W.}},
\bauthor{\bsnm{Liu}, \binits{L.}},
\bauthor{\bsnm{Long}, \binits{G.}}:
\bctitle{Interpretable time-series classification on few-shot samples}.
In: \bbtitle{2020 International Joint Conference on Neural Networks (IJCNN)},
pp. \bfpage{1}--\blpage{8}
(\byear{2020}).
\bcomment{IEEE}
\end{bchapter}
\endbibitem

\bibitem[\protect\citeauthoryear{Narwariya et~al.}{2020}]{narwariya2020meta}
\begin{bchapter}
\bauthor{\bsnm{Narwariya}, \binits{J.}},
\bauthor{\bsnm{Malhotra}, \binits{P.}},
\bauthor{\bsnm{Vig}, \binits{L.}},
\bauthor{\bsnm{Shroff}, \binits{G.}},
\bauthor{\bsnm{Vishnu}, \binits{T.}}:
\bctitle{Meta-learning for few-shot time series classification}.
In: \bbtitle{Proceedings of the 7th ACM IKDD CoDS and 25th COMAD},
pp. \bfpage{28}--\blpage{36}
(\byear{2020})
\end{bchapter}
\endbibitem

\bibitem[\protect\citeauthoryear{Wang et~al.}{2020}]{wang2020generalizing}
\begin{barticle}
\bauthor{\bsnm{Wang}, \binits{Y.}},
\bauthor{\bsnm{Yao}, \binits{Q.}},
\bauthor{\bsnm{Kwok}, \binits{J.T.}},
\bauthor{\bsnm{Ni}, \binits{L.M.}}:
\batitle{Generalizing from a few examples: A survey on few-shot learning}.
\bjtitle{ACM computing surveys (csur)}
\bvolume{53}(\bissue{3}),
\bfpage{1}--\blpage{34}
(\byear{2020})
\end{barticle}
\endbibitem

\bibitem[\protect\citeauthoryear{Jin et~al.}{2024}]{jin2024position}
\begin{bchapter}
\bauthor{\bsnm{Jin}, \binits{M.}},
\bauthor{\bsnm{Zhang}, \binits{Y.}},
\bauthor{\bsnm{Chen}, \binits{W.}},
\bauthor{\bsnm{Zhang}, \binits{K.}},
\bauthor{\bsnm{Liang}, \binits{Y.}},
\bauthor{\bsnm{Yang}, \binits{B.}},
\bauthor{\bsnm{Wang}, \binits{J.}},
\bauthor{\bsnm{Pan}, \binits{S.}},
\bauthor{\bsnm{Wen}, \binits{Q.}}:
\bctitle{Position: What can large language models tell us about time series analysis}.
In: \bbtitle{Forty-first International Conference on Machine Learning}
(\byear{2024})
\end{bchapter}
\endbibitem

\bibitem[\protect\citeauthoryear{Gudmundsson et~al.}{2008}]{gudmundsson2008support}
\begin{bchapter}
\bauthor{\bsnm{Gudmundsson}, \binits{S.}},
\bauthor{\bsnm{Runarsson}, \binits{T.P.}},
\bauthor{\bsnm{Sigurdsson}, \binits{S.}}:
\bctitle{Support vector machines and dynamic time warping for time series}.
In: \bbtitle{2008 IEEE International Joint Conference on Neural Networks (IEEE World Congress on Computational Intelligence)},
pp. \bfpage{2772}--\blpage{2776}
(\byear{2008}).
\bcomment{IEEE}
\end{bchapter}
\endbibitem

\bibitem[\protect\citeauthoryear{Dempster et~al.}{2020}]{dempster2020rocket}
\begin{barticle}
\bauthor{\bsnm{Dempster}, \binits{A.}},
\bauthor{\bsnm{Petitjean}, \binits{F.}},
\bauthor{\bsnm{Webb}, \binits{G.I.}}:
\batitle{Rocket: exceptionally fast and accurate time series classification using random convolutional kernels}.
\bjtitle{Data Mining and Knowledge Discovery}
\bvolume{34}(\bissue{5}),
\bfpage{1454}--\blpage{1495}
(\byear{2020})
\end{barticle}
\endbibitem

\bibitem[\protect\citeauthoryear{Karim et~al.}{2017}]{karim2017lstm}
\begin{barticle}
\bauthor{\bsnm{Karim}, \binits{F.}},
\bauthor{\bsnm{Majumdar}, \binits{S.}},
\bauthor{\bsnm{Darabi}, \binits{H.}},
\bauthor{\bsnm{Chen}, \binits{S.}}:
\batitle{Lstm fully convolutional networks for time series classification}.
\bjtitle{IEEE access}
\bvolume{6},
\bfpage{1662}--\blpage{1669}
(\byear{2017})
\end{barticle}
\endbibitem

\bibitem[\protect\citeauthoryear{Tang et~al.}{2020}]{tang2020rethinking}
\begin{botherref}
\oauthor{\bsnm{Tang}, \binits{W.}},
\oauthor{\bsnm{Long}, \binits{G.}},
\oauthor{\bsnm{Liu}, \binits{L.}},
\oauthor{\bsnm{Zhou}, \binits{T.}},
\oauthor{\bsnm{Jiang}, \binits{J.}},
\oauthor{\bsnm{Blumenstein}, \binits{M.}}:
Rethinking 1d-cnn for time series classification: A stronger baseline.
arXiv preprint arXiv:2002.10061,
1--7
(2020)
\end{botherref}
\endbibitem

\bibitem[\protect\citeauthoryear{Yang et~al.}{2024}]{yang2024dyformer}
\begin{barticle}
\bauthor{\bsnm{Yang}, \binits{C.}},
\bauthor{\bsnm{Wang}, \binits{X.}},
\bauthor{\bsnm{Yao}, \binits{L.}},
\bauthor{\bsnm{Long}, \binits{G.}},
\bauthor{\bsnm{Xu}, \binits{G.}}:
\batitle{Dyformer: A dynamic transformer-based architecture for multivariate time series classification}.
\bjtitle{Information Sciences}
\bvolume{656},
\bfpage{119881}
(\byear{2024})
\end{barticle}
\endbibitem

\bibitem[\protect\citeauthoryear{Le et~al.}{2024}]{le2024shapeformer}
\begin{botherref}
\oauthor{\bsnm{Le}, \binits{X.-M.}},
\oauthor{\bsnm{Luo}, \binits{L.}},
\oauthor{\bsnm{Aickelin}, \binits{U.}},
\oauthor{\bsnm{Tran}, \binits{M.-T.}}:
Shapeformer: Shapelet transformer for multivariate time series classification.
arXiv preprint arXiv:2405.14608
(2024)
\end{botherref}
\endbibitem

\bibitem[\protect\citeauthoryear{Gupta et~al.}{2021}]{gupta2021similarity}
\begin{bchapter}
\bauthor{\bsnm{Gupta}, \binits{P.}},
\bauthor{\bsnm{Bhaskarpandit}, \binits{S.}},
\bauthor{\bsnm{Gupta}, \binits{M.}}:
\bctitle{Similarity learning based few shot learning for ecg time series classification}.
In: \bbtitle{2021 Digital Image Computing: Techniques and Applications (DICTA)},
pp. \bfpage{1}--\blpage{8}
(\byear{2021}).
\bcomment{IEEE}
\end{bchapter}
\endbibitem

\bibitem[\protect\citeauthoryear{Zhang et~al.}{2023}]{zhang2023few}
\begin{bchapter}
\bauthor{\bsnm{Zhang}, \binits{H.}},
\bauthor{\bsnm{Pang}, \binits{Z.}},
\bauthor{\bsnm{Wang}, \binits{J.}},
\bauthor{\bsnm{Li}, \binits{T.}}:
\bctitle{Few-shot learning using data augmentation and time-frequency transformation for time series classification}.
In: \bbtitle{2023 5th International Conference on Robotics, Intelligent Control and Artificial Intelligence (RICAI)},
pp. \bfpage{733}--\blpage{738}
(\byear{2023}).
\bcomment{IEEE}
\end{bchapter}
\endbibitem

\bibitem[\protect\citeauthoryear{Park et~al.}{2023}]{park2023meta}
\begin{barticle}
\bauthor{\bsnm{Park}, \binits{S.-H.}},
\bauthor{\bsnm{Syazwany}, \binits{N.S.}},
\bauthor{\bsnm{Lee}, \binits{S.-C.}}:
\batitle{Meta-feature fusion for few-shot time series classification}.
\bjtitle{IEEE Access}
\bvolume{11},
\bfpage{41400}--\blpage{41414}
(\byear{2023})
\end{barticle}
\endbibitem

\bibitem[\protect\citeauthoryear{Chen et~al.}{2023}]{chen2023supervised}
\begin{bchapter}
\bauthor{\bsnm{Chen}, \binits{X.}},
\bauthor{\bsnm{Ge}, \binits{C.}},
\bauthor{\bsnm{Wang}, \binits{M.}},
\bauthor{\bsnm{Wang}, \binits{J.}}:
\bctitle{Supervised contrastive few-shot learning for high-frequency time series}.
In: \bbtitle{Proceedings of the AAAI Conference on Artificial Intelligence},
vol. \bseriesno{37},
pp. \bfpage{7069}--\blpage{7077}
(\byear{2023})
\end{bchapter}
\endbibitem

\bibitem[\protect\citeauthoryear{Gruver et~al.}{2024}]{gruver2024large}
\begin{botherref}
\oauthor{\bsnm{Gruver}, \binits{N.}},
\oauthor{\bsnm{Finzi}, \binits{M.}},
\oauthor{\bsnm{Qiu}, \binits{S.}},
\oauthor{\bsnm{Wilson}, \binits{A.G.}}:
Large language models are zero-shot time series forecasters.
Advances in Neural Information Processing Systems
\textbf{36}
(2024)
\end{botherref}
\endbibitem

\bibitem[\protect\citeauthoryear{Xue and Salim}{2023}]{xue2023promptcast}
\begin{botherref}
\oauthor{\bsnm{Xue}, \binits{H.}},
\oauthor{\bsnm{Salim}, \binits{F.D.}}:
Promptcast: A new prompt-based learning paradigm for time series forecasting.
IEEE Transactions on Knowledge and Data Engineering
(2023)
\end{botherref}
\endbibitem

\bibitem[\protect\citeauthoryear{Jin et~al.}{2023}]{jin2023time}
\begin{botherref}
\oauthor{\bsnm{Jin}, \binits{M.}},
\oauthor{\bsnm{Wang}, \binits{S.}},
\oauthor{\bsnm{Ma}, \binits{L.}},
\oauthor{\bsnm{Chu}, \binits{Z.}},
\oauthor{\bsnm{Zhang}, \binits{J.Y.}},
\oauthor{\bsnm{Shi}, \binits{X.}},
\oauthor{\bsnm{Chen}, \binits{P.-Y.}},
\oauthor{\bsnm{Liang}, \binits{Y.}},
\oauthor{\bsnm{Li}, \binits{Y.-F.}},
\oauthor{\bsnm{Pan}, \binits{S.}}, et al.:
Time-llm: Time series forecasting by reprogramming large language models.
arXiv preprint arXiv:2310.01728
(2023)
\end{botherref}
\endbibitem

\bibitem[\protect\citeauthoryear{Zhou et~al.}{2023}]{zhou2023one}
\begin{barticle}
\bauthor{\bsnm{Zhou}, \binits{T.}},
\bauthor{\bsnm{Niu}, \binits{P.}},
\bauthor{\bsnm{Sun}, \binits{L.}},
\bauthor{\bsnm{Jin}, \binits{R.}}, \betal:
\batitle{One fits all: Power general time series analysis by pretrained lm}.
\bjtitle{Advances in neural information processing systems}
\bvolume{36},
\bfpage{43322}--\blpage{43355}
(\byear{2023})
\end{barticle}
\endbibitem

\bibitem[\protect\citeauthoryear{Cao et~al.}{2023}]{cao2023tempo}
\begin{botherref}
\oauthor{\bsnm{Cao}, \binits{D.}},
\oauthor{\bsnm{Jia}, \binits{F.}},
\oauthor{\bsnm{Arik}, \binits{S.O.}},
\oauthor{\bsnm{Pfister}, \binits{T.}},
\oauthor{\bsnm{Zheng}, \binits{Y.}},
\oauthor{\bsnm{Ye}, \binits{W.}},
\oauthor{\bsnm{Liu}, \binits{Y.}}:
Tempo: Prompt-based generative pre-trained transformer for time series forecasting.
arXiv preprint arXiv:2310.04948
(2023)
\end{botherref}
\endbibitem

\bibitem[\protect\citeauthoryear{Pan et~al.}{2024}]{pan2024s}
\begin{bchapter}
\bauthor{\bsnm{Pan}, \binits{Z.}},
\bauthor{\bsnm{Jiang}, \binits{Y.}},
\bauthor{\bsnm{Garg}, \binits{S.}},
\bauthor{\bsnm{Schneider}, \binits{A.}},
\bauthor{\bsnm{Nevmyvaka}, \binits{Y.}},
\bauthor{\bsnm{Song}, \binits{D.}}:
\bctitle{$\textbf{S}^2$ip-llm: Semantic space informed prompt learning with llm for time series forecasting}.
In: \bbtitle{Forty-first International Conference on Machine Learning}
(\byear{2024})
\end{bchapter}
\endbibitem

\bibitem[\protect\citeauthoryear{Nie et~al.}{2022}]{nie2022time}
\begin{botherref}
\oauthor{\bsnm{Nie}, \binits{Y.}},
\oauthor{\bsnm{Nguyen}, \binits{N.H.}},
\oauthor{\bsnm{Sinthong}, \binits{P.}},
\oauthor{\bsnm{Kalagnanam}, \binits{J.}}:
A time series is worth 64 words: Long-term forecasting with transformers.
arXiv preprint arXiv:2211.14730
(2022)
\end{botherref}
\endbibitem

\bibitem[\protect\citeauthoryear{Bai et~al.}{2018}]{bai2018empirical}
\begin{botherref}
\oauthor{\bsnm{Bai}, \binits{S.}},
\oauthor{\bsnm{Kolter}, \binits{J.Z.}},
\oauthor{\bsnm{Koltun}, \binits{V.}}:
An empirical evaluation of generic convolutional and recurrent networks for sequence modeling.
arXiv preprint arXiv:1803.01271
(2018)
\end{botherref}
\endbibitem

\bibitem[\protect\citeauthoryear{Hu et~al.}{2021}]{hu2021lora}
\begin{botherref}
\oauthor{\bsnm{Hu}, \binits{E.J.}},
\oauthor{\bsnm{Shen}, \binits{Y.}},
\oauthor{\bsnm{Wallis}, \binits{P.}},
\oauthor{\bsnm{Allen-Zhu}, \binits{Z.}},
\oauthor{\bsnm{Li}, \binits{Y.}},
\oauthor{\bsnm{Wang}, \binits{S.}},
\oauthor{\bsnm{Wang}, \binits{L.}},
\oauthor{\bsnm{Chen}, \binits{W.}}:
Lora: Low-rank adaptation of large language models.
arXiv preprint arXiv:2106.09685
(2021)
\end{botherref}
\endbibitem

\bibitem[\protect\citeauthoryear{Bagnall et~al.}{2018}]{bagnall2018uea}
\begin{botherref}
\oauthor{\bsnm{Bagnall}, \binits{A.}},
\oauthor{\bsnm{Dau}, \binits{H.A.}},
\oauthor{\bsnm{Lines}, \binits{J.}},
\oauthor{\bsnm{Flynn}, \binits{M.}},
\oauthor{\bsnm{Large}, \binits{J.}},
\oauthor{\bsnm{Bostrom}, \binits{A.}},
\oauthor{\bsnm{Southam}, \binits{P.}},
\oauthor{\bsnm{Keogh}, \binits{E.}}:
The uea multivariate time series classification archive, 2018.
arXiv preprint arXiv:1811.00075
(2018)
\end{botherref}
\endbibitem

\bibitem[\protect\citeauthoryear{Zeng et~al.}{2023}]{zeng2023transformers}
\begin{bchapter}
\bauthor{\bsnm{Zeng}, \binits{A.}},
\bauthor{\bsnm{Chen}, \binits{M.}},
\bauthor{\bsnm{Zhang}, \binits{L.}},
\bauthor{\bsnm{Xu}, \binits{Q.}}:
\bctitle{Are transformers effective for time series forecasting?}
In: \bbtitle{Proceedings of the AAAI Conference on Artificial Intelligence},
vol. \bseriesno{37},
pp. \bfpage{11121}--\blpage{11128}
(\byear{2023})
\end{bchapter}
\endbibitem

\bibitem[\protect\citeauthoryear{Wu et~al.}{2022}]{wu2022timesnet}
\begin{botherref}
\oauthor{\bsnm{Wu}, \binits{H.}},
\oauthor{\bsnm{Hu}, \binits{T.}},
\oauthor{\bsnm{Liu}, \binits{Y.}},
\oauthor{\bsnm{Zhou}, \binits{H.}},
\oauthor{\bsnm{Wang}, \binits{J.}},
\oauthor{\bsnm{Long}, \binits{M.}}:
Timesnet: Temporal 2d-variation modeling for general time series analysis.
arXiv preprint arXiv:2210.02186
(2022)
\end{botherref}
\endbibitem

\bibitem[\protect\citeauthoryear{Wu et~al.}{2021}]{wu2021autoformer}
\begin{barticle}
\bauthor{\bsnm{Wu}, \binits{H.}},
\bauthor{\bsnm{Xu}, \binits{J.}},
\bauthor{\bsnm{Wang}, \binits{J.}},
\bauthor{\bsnm{Long}, \binits{M.}}:
\batitle{Autoformer: Decomposition transformers with auto-correlation for long-term series forecasting}.
\bjtitle{Advances in neural information processing systems}
\bvolume{34},
\bfpage{22419}--\blpage{22430}
(\byear{2021})
\end{barticle}
\endbibitem

\bibitem[\protect\citeauthoryear{Zhang and Yan}{2023}]{zhang2023crossformer}
\begin{bchapter}
\bauthor{\bsnm{Zhang}, \binits{Y.}},
\bauthor{\bsnm{Yan}, \binits{J.}}:
\bctitle{Crossformer: Transformer utilizing cross-dimension dependency for multivariate time series forecasting}.
In: \bbtitle{The Eleventh International Conference on Learning Representations}
(\byear{2023})
\end{bchapter}
\endbibitem

\bibitem[\protect\citeauthoryear{Zhou et~al.}{2022}]{zhou2022fedformer}
\begin{bchapter}
\bauthor{\bsnm{Zhou}, \binits{T.}},
\bauthor{\bsnm{Ma}, \binits{Z.}},
\bauthor{\bsnm{Wen}, \binits{Q.}},
\bauthor{\bsnm{Wang}, \binits{X.}},
\bauthor{\bsnm{Sun}, \binits{L.}},
\bauthor{\bsnm{Jin}, \binits{R.}}:
\bctitle{Fedformer: Frequency enhanced decomposed transformer for long-term series forecasting}.
In: \bbtitle{International Conference on Machine Learning},
pp. \bfpage{27268}--\blpage{27286}
(\byear{2022}).
\bcomment{PMLR}
\end{bchapter}
\endbibitem

\bibitem[\protect\citeauthoryear{Zhou et~al.}{2021}]{zhou2021informer}
\begin{bchapter}
\bauthor{\bsnm{Zhou}, \binits{H.}},
\bauthor{\bsnm{Zhang}, \binits{S.}},
\bauthor{\bsnm{Peng}, \binits{J.}},
\bauthor{\bsnm{Zhang}, \binits{S.}},
\bauthor{\bsnm{Li}, \binits{J.}},
\bauthor{\bsnm{Xiong}, \binits{H.}},
\bauthor{\bsnm{Zhang}, \binits{W.}}:
\bctitle{Informer: Beyond efficient transformer for long sequence time-series forecasting}.
In: \bbtitle{Proceedings of the AAAI Conference on Artificial Intelligence},
vol. \bseriesno{35},
pp. \bfpage{11106}--\blpage{11115}
(\byear{2021})
\end{bchapter}
\endbibitem

\bibitem[\protect\citeauthoryear{Kitaev et~al.}{2020}]{kitaev2020reformer}
\begin{botherref}
\oauthor{\bsnm{Kitaev}, \binits{N.}},
\oauthor{\bsnm{Kaiser}, \binits{{\L}.}},
\oauthor{\bsnm{Levskaya}, \binits{A.}}:
Reformer: The efficient transformer.
arXiv preprint arXiv:2001.04451
(2020)
\end{botherref}
\endbibitem

\bibitem[\protect\citeauthoryear{Vaswani et~al.}{2017}]{vaswani2017attention}
\begin{botherref}
\oauthor{\bsnm{Vaswani}, \binits{A.}},
\oauthor{\bsnm{Shazeer}, \binits{N.}},
\oauthor{\bsnm{Parmar}, \binits{N.}},
\oauthor{\bsnm{Uszkoreit}, \binits{J.}},
\oauthor{\bsnm{Jones}, \binits{L.}},
\oauthor{\bsnm{Gomez}, \binits{A.N.}},
\oauthor{\bsnm{Kaiser}, \binits{{\L}.}},
\oauthor{\bsnm{Polosukhin}, \binits{I.}}:
Attention is all you need.
Advances in neural information processing systems
\textbf{30}
(2017)
\end{botherref}
\endbibitem

\bibitem[\protect\citeauthoryear{Radford et~al.}{2019}]{radford2019language}
\begin{barticle}
\bauthor{\bsnm{Radford}, \binits{A.}},
\bauthor{\bsnm{Wu}, \binits{J.}},
\bauthor{\bsnm{Child}, \binits{R.}},
\bauthor{\bsnm{Luan}, \binits{D.}},
\bauthor{\bsnm{Amodei}, \binits{D.}},
\bauthor{\bsnm{Sutskever}, \binits{I.}}, \betal:
\batitle{Language models are unsupervised multitask learners}.
\bjtitle{OpenAI blog}
\bvolume{1}(\bissue{8}),
\bfpage{9}
(\byear{2019})
\end{barticle}
\endbibitem

\bibitem[\protect\citeauthoryear{Bai et~al.}{2023}]{bai2023qwen}
\begin{botherref}
\oauthor{\bsnm{Bai}, \binits{J.}},
\oauthor{\bsnm{Bai}, \binits{S.}},
\oauthor{\bsnm{Chu}, \binits{Y.}},
\oauthor{\bsnm{Cui}, \binits{Z.}},
\oauthor{\bsnm{Dang}, \binits{K.}},
\oauthor{\bsnm{Deng}, \binits{X.}},
\oauthor{\bsnm{Fan}, \binits{Y.}},
\oauthor{\bsnm{Ge}, \binits{W.}},
\oauthor{\bsnm{Han}, \binits{Y.}},
\oauthor{\bsnm{Huang}, \binits{F.}}, et al.:
Qwen technical report.
arXiv preprint arXiv:2309.16609
(2023)
\end{botherref}
\endbibitem

\bibitem[\protect\citeauthoryear{Abdin et~al.}{2024}]{abdin2024phi}
\begin{botherref}
\oauthor{\bsnm{Abdin}, \binits{M.}},
\oauthor{\bsnm{Jacobs}, \binits{S.A.}},
\oauthor{\bsnm{Awan}, \binits{A.A.}},
\oauthor{\bsnm{Aneja}, \binits{J.}},
\oauthor{\bsnm{Awadallah}, \binits{A.}},
\oauthor{\bsnm{Awadalla}, \binits{H.}},
\oauthor{\bsnm{Bach}, \binits{N.}},
\oauthor{\bsnm{Bahree}, \binits{A.}},
\oauthor{\bsnm{Bakhtiari}, \binits{A.}},
\oauthor{\bsnm{Behl}, \binits{H.}}, et al.:
Phi-3 technical report: A highly capable language model locally on your phone.
arXiv preprint arXiv:2404.14219
(2024)
\end{botherref}
\endbibitem

\bibitem[\protect\citeauthoryear{Dubey et~al.}{2024}]{dubey2024llama}
\begin{botherref}
\oauthor{\bsnm{Dubey}, \binits{A.}},
\oauthor{\bsnm{Jauhri}, \binits{A.}},
\oauthor{\bsnm{Pandey}, \binits{A.}},
\oauthor{\bsnm{Kadian}, \binits{A.}},
\oauthor{\bsnm{Al-Dahle}, \binits{A.}},
\oauthor{\bsnm{Letman}, \binits{A.}},
\oauthor{\bsnm{Mathur}, \binits{A.}},
\oauthor{\bsnm{Schelten}, \binits{A.}},
\oauthor{\bsnm{Yang}, \binits{A.}},
\oauthor{\bsnm{Fan}, \binits{A.}}, et al.:
The llama 3 herd of models.
arXiv preprint arXiv:2407.21783
(2024)
\end{botherref}
\endbibitem

\bibitem[\protect\citeauthoryear{Biderman et~al.}{2023}]{biderman2023pythia}
\begin{bchapter}
\bauthor{\bsnm{Biderman}, \binits{S.}},
\bauthor{\bsnm{Schoelkopf}, \binits{H.}},
\bauthor{\bsnm{Anthony}, \binits{Q.G.}},
\bauthor{\bsnm{Bradley}, \binits{H.}},
\bauthor{\bsnm{O’Brien}, \binits{K.}},
\bauthor{\bsnm{Hallahan}, \binits{E.}},
\bauthor{\bsnm{Khan}, \binits{M.A.}},
\bauthor{\bsnm{Purohit}, \binits{S.}},
\bauthor{\bsnm{Prashanth}, \binits{U.S.}},
\bauthor{\bsnm{Raff}, \binits{E.}}, \betal:
\bctitle{Pythia: A suite for analyzing large language models across training and scaling}.
In: \bbtitle{International Conference on Machine Learning},
pp. \bfpage{2397}--\blpage{2430}
(\byear{2023}).
\bcomment{PMLR}
\end{bchapter}
\endbibitem

\bibitem[\protect\citeauthoryear{Lu et~al.}{2021}]{lu2021inter}
\begin{botherref}
\oauthor{\bsnm{Lu}, \binits{P.}},
\oauthor{\bsnm{Gong}, \binits{R.}},
\oauthor{\bsnm{Jiang}, \binits{S.}},
\oauthor{\bsnm{Qiu}, \binits{L.}},
\oauthor{\bsnm{Huang}, \binits{S.}},
\oauthor{\bsnm{Liang}, \binits{X.}},
\oauthor{\bsnm{Zhu}, \binits{S.-C.}}:
Inter-gps: Interpretable geometry problem solving with formal language and symbolic reasoning.
arXiv preprint arXiv:2105.04165
(2021)
\end{botherref}
\endbibitem

\end{thebibliography}

\end{document}